\documentclass{opendatalab}

\usepackage[utf8]{inputenc}
\usepackage{xcolor}
\usepackage[most]{tcolorbox}
\usepackage{longtable}
\usepackage{listings}
\usepackage[numbers]{natbib}
\usepackage{enumitem}
\usepackage{graphicx}
\usepackage{xspace} 
\usepackage{hyperref}
\usepackage{pifont}

\definecolor{codegreen}{rgb}{0,0.6,0}
\definecolor{codegray}{rgb}{0.5,0.5,0.5}
\definecolor{codepurple}{rgb}{0.58,0,0.82}
\definecolor{backcolour}{rgb}{0.95,0.95,0.92}
\definecolor{promptcolor}{HTML}{D1D0F2}
\definecolor{promptcolorheader}{HTML}{bdbcec}
\usepackage[table]{xcolor}
\newcommand{\github}{\raisebox{-1.5pt}{\includegraphics[height=1.05em]{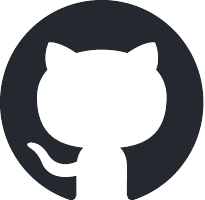}}\xspace}
\newcommand{\web}{\raisebox{-1.5pt}{\includegraphics[height=1.05em]{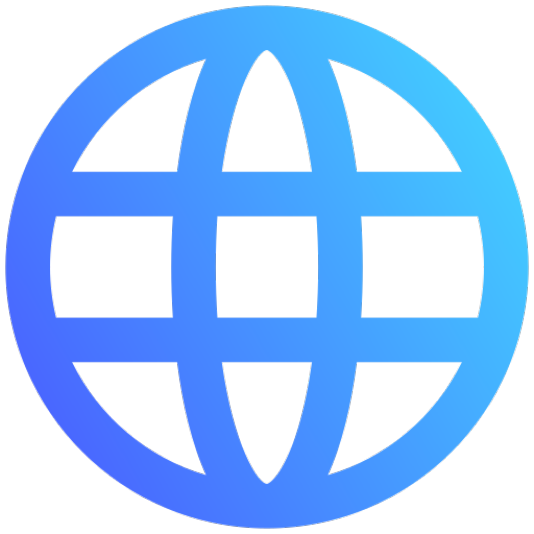}}\xspace}
\newcommand{\huggingface}{\raisebox{-1.5pt}{\includegraphics[height=1.05em]{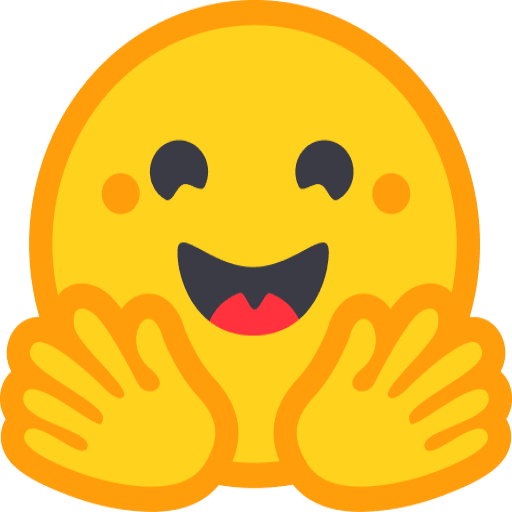}}\xspace}

\definecolor{promptcolor}{HTML}{D1D0F2}
\definecolor{promptcolorheader}{HTML}{bdbcec}

\newtcolorbox{promptbox}[1][]{
  enhanced, breakable,
  top=0.3em,bottom=0.3em,left=0.5em,right=0.5em,
  toptitle=0.3em,bottomtitle=0.2em,boxsep=0pt,
  colframe=promptcolorheader, colback=promptcolor!50, boxrule=0.5pt,
  width=\columnwidth,        
  title={\footnotesize #1}   
}
\lstdefinestyle{mystyle}{
    backgroundcolor=\color{backcolour},   
    commentstyle=\color{codegreen},
    keywordstyle=\color{magenta},
    numberstyle=\tiny\color{codegray},
    stringstyle=\color{codepurple},
    basicstyle=\ttfamily\footnotesize,
    breakatwhitespace=false,         
    breaklines=true,                 
    captionpos=b,                    
    keepspaces=true,                 
    numbers=left,                    
    numbersep=5pt,                  
    showspaces=false,                
    showstringspaces=false,
    showtabs=false,                  
    tabsize=2
}
\usepackage{enumitem}
\setlist[itemize]{leftmargin=*, itemsep=1pt}

\title{GGBench: A Geometric Generative Reasoning Benchmark for Unified Multimodal Models}

\author[1]{Jingxuan Wei$^\dagger$}
\author[1]{Caijun Jia$^\dagger$}
\author[1]{Xi Bai$^\dagger$}
\author[1]{Xinglong Xu$^\dagger$}
\author[2]{Siyuan Li}
\author[1]{Linzhuang Sun}
\author[1]{Bihui Yu}
\author[2]{Conghui He}
\author[2]{Lijun Wu}
\author[2]{Cheng Tan}

\affiliation[1]{University of Chinese Academy of Sciences}
\affiliation[2]{Shanghai Artificial Intelligence Laboratory}

\abstract{
\vspace{-4mm}
The advent of Unified Multimodal Models (UMMs) signals a paradigm shift in artificial intelligence, moving from passive perception to active, cross-modal generation. Despite their unprecedented ability to synthesize information, a critical gap persists in evaluation: existing benchmarks primarily assess discriminative understanding or unconstrained image generation separately, failing to measure the integrated cognitive process of generative reasoning. To bridge this gap, we propose that geometric construction provides an ideal testbed as it inherently demands a fusion of language comprehension and precise visual generation. We introduce GGBench, a benchmark designed specifically to evaluate geometric generative reasoning. It provides a comprehensive framework for systematically diagnosing a model's ability to not only understand and reason but to actively construct a solution, thereby setting a more rigorous standard for the next generation of intelligent systems.
\vspace{-2mm}
}

\date{\today}
\correspondence{Cheng Tan, \email{tancheng@pjlab.org.cn}
\\
  \parbox{\linewidth}{\centering
    \github~\href{https://github.com/openraiser/GGBench}{\textbf{Code}} \quad
    \web~\href{https://openraiser.github.io/GGBench}{\textbf{Website}} \quad
\huggingface~\href{https://huggingface.co/datasets/openraiser/GGBench}{\textbf{Dataset}}
  }
  \vspace{-6mm}
}

\begin{document}
\maketitle

\begin{figure}[h]
  \centering
  \vspace{-2mm}
  \includegraphics[width=0.96\linewidth]{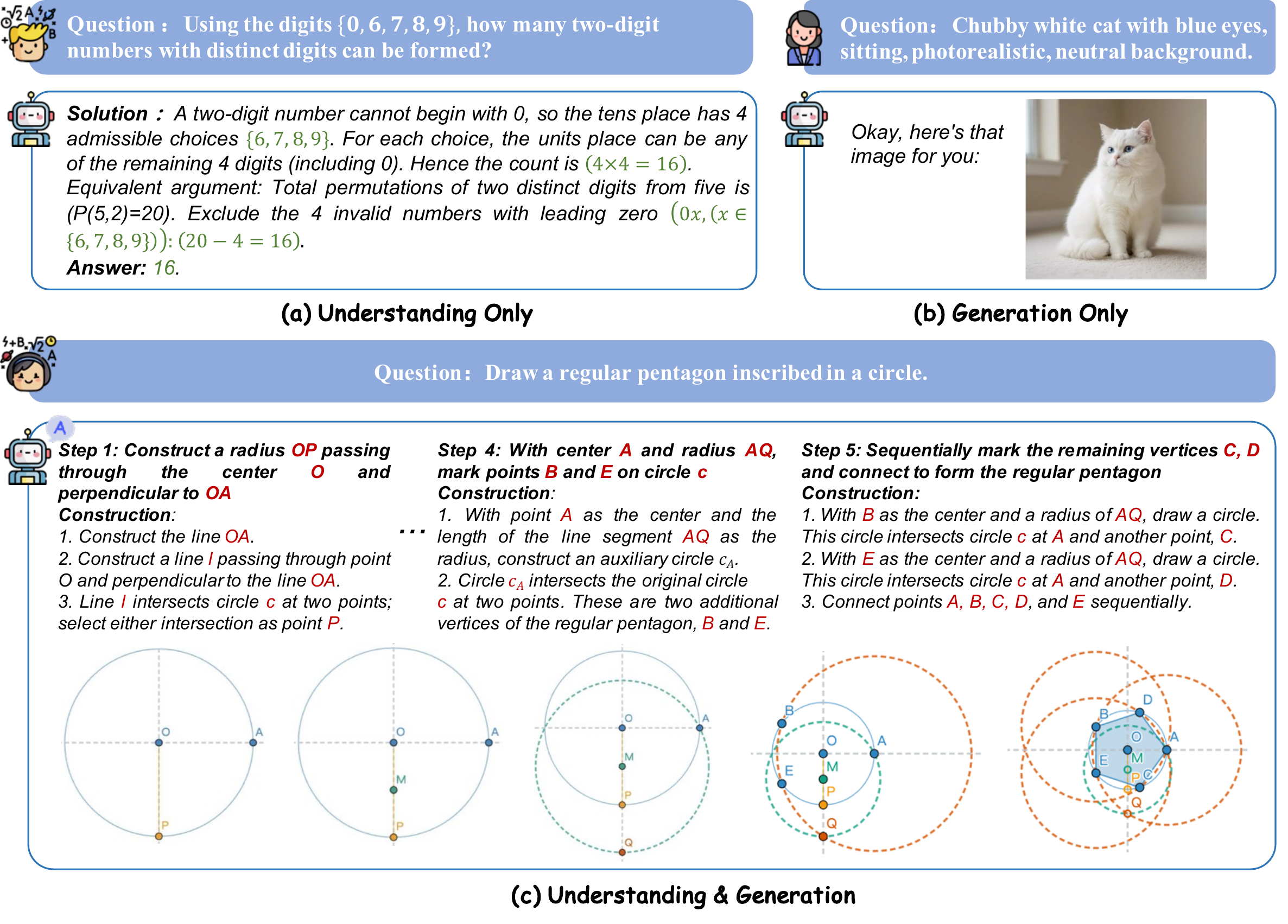}
  \vspace{-4mm}
  \caption{The paradigm shift to generative reasoning. Conventional benchmarks evaluate (a) Understanding or (b) Generation in isolation. GGBench introduces (c) integrated Understanding\& Generation evaluation, requiring generative reasoning from Unified Multimodal Models.}
  \vspace{-4mm}
  \label{fig:teaser}
\end{figure}

\clearpage

\section{Introduction}
\label{section:intro}

Artificial intelligence research is undergoing a profound paradigm shift, evolving from the foundational era of passive perception to the more sophisticated stage of contextual understanding~\cite{alayrac2022flamingo}, and now culminating in the frontier of active generation. This evolutionary leap is epitomized by the recent advent of Unified Multimodal Models (UMMs)\cite{zhang2025unified}, such as the groundbreaking GPT-4o\cite{hurst2024gpt} and Nano Banana~\cite{google_gemini25_flash_image_2025}, which demonstrate an unprecedented ability to fluently process, reason across, and synthesize information from a diverse spectrum of modalities. These UMMs can not only understand complex visual and textual inputs, but also generate rich outputs; notably, advanced systems like Unified-IO 2~\cite{lu2024unified} can even produce images, audio, or actions in response to multimodal instructions. Despite these breakthroughs, existing evaluation benchmarks lag behind – they primarily assess discriminative reasoning (e.g. selecting an answer or classifying a visual input) and thus fail to capture the generative dimension of intelligence~\cite{mmeu}. In particular, current tests rarely challenge models to construct solutions (e.g. draw a diagram or formulate a step-by-step proof), which is often essential for domains like geometry. This gap is especially evident in geometric problem-solving, where genuine understanding is inseparable from the ability to plan and generate a diagram through multi-step spatial reasoning~\cite{wei2025geoint}. Indeed, current models struggle with this constructive aspect – even advanced multimodal systems have difficulty reliably interpreting complex geometric setups~\cite{gllava,trinh2024alphageometry,duan2025codeplot}.

The trajectory of evaluation benchmarks over recent years reflects a push toward increasingly complex cognition, yet also reveals a gap in evaluating \textbf{generative reasoning}. Early benchmarks targeted single-modality reasoning or generation, primarily for Large Language Models (LLMs) in purely textual domains. For example, GSM8K~\cite{cobbe2021training} and MATH~\cite{hendrycks2021measuring} established rigorous tests for mathematical problem-solving in text, requiring models to parse word problems and generate final answers through a step-by-step reasoning chain. Techniques such as chain-of-thought prompting greatly improved performance on these datasets~\cite{wei2022chain}. Building on this progress, the community introduced multimodal reasoning challenges that incorporate visual context. Benchmarks like ScienceQA~\cite{lu2022learn} require models to combine text understanding with image comprehension~\cite{tan2024boosting}, and the recent MathVista suite~\cite{mathvista} amalgamates 28 math and visual datasets to evaluate mathematical reasoning in visual contexts. These tasks, aimed at Multimodal LLMs (MLLMs), represent a critical step toward contextual understanding – models must interpret a question against an image or figure – but they remain largely discriminative. The model is still to answer or classify based on given content, rather than to create new content. The latest multimodal benchmarks have started to probe both understanding and generation capabilities, but often in a decoupled fashion. Efforts like MME~\cite{zhang2021mme}, MM-Vet~\cite{yu2023mm}, and MMBench~\cite{liu2024mmbench} evaluate a broad range of perceptual and cognitive tasks to score models on each ability individually. Critically, however, even these comprehensive benchmarks treat understanding and generation as separate modules. This fragmentation means we lack evaluation of the integrated cognitive process – the scenario where a system must simultaneously comprehend, reason, and generate a complex outcome.

\begin{figure}[h]
  \centering
  \includegraphics[width=\linewidth]{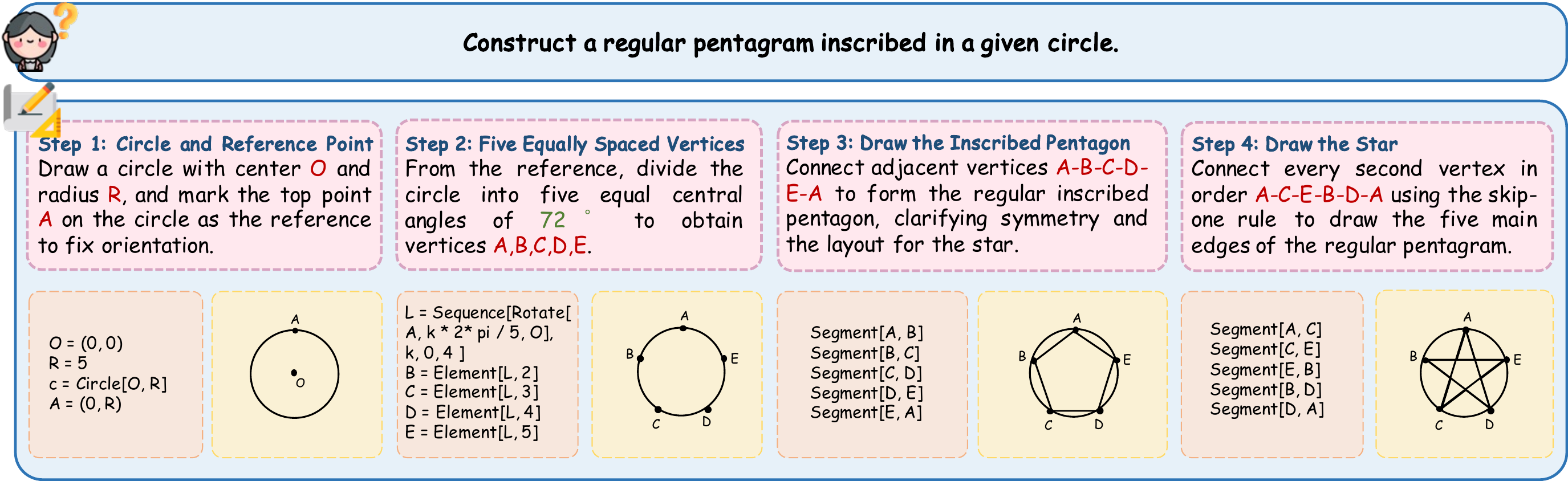}
  \vspace{-4mm}
  \caption{GGBench's step-by-step evaluation. Beyond traditional text-image pairs, GGBench provides executable code for each construction step, allowing for precise and automated verification.}
  \vspace{-3mm}
  \label{fig:tri-modal}
\end{figure}

Perform \textbf{generative reasoning} demands both cross-modal understanding of abstract concepts and the structured, stepwise generation of a coherent solution. We propose that geometric construction problems provide an ideal testbed for evaluating such integrated understanding–generation capabilities within a unified setting. Such problems inherently require a model to: 
\begin{itemize}[leftmargin=2em]
    \vspace{-1mm}
    \item Parse and comprehend abstract natural language instructions with domain-specific constraints;
    \vspace{-1mm}
    \item Formulate a multi-step plan grounded in formal geometric principles;
    \vspace{-1mm}
    \item Generating precise figures that satisfies the given constraints.
    \vspace{-1mm}
\end{itemize}
Performance on geometric construction tasks is directly verifiable, offering a clear and objective criterion for evaluation. Success in such tasks engages the full spectrum of intelligence: language understanding, mathematical reasoning, long-horizon planning, and visual generation. To enable this rigorous evaluation, we introduce GGBench, a comprehensive benchmark where each problem instance consists of precisely aligned text, code, and image modalities, as shown in Figure~\ref{fig:tri-modal}, facilitating a systematic analysis of the entire multimodal generative reasoning process.

\section{Related Work}
\subsection{The Evolution of Mathematical Reasoning Benchmarks}

The evaluation of mathematical reasoning in AI has progressed from discriminative question answering toward more generative, process-oriented tasks. Early benchmarks such as GeoQA~\cite{chen2021geoqa} provides valuable testbeds but narrowly targets plane figures and evaluates models on answer accuracy. Several recent benchmarks like MATH-Vision (MATH-V)~\cite{mathv}, MathVista~\cite{mathvista}, and MathVerse~\cite{mathverse} have broadened the scope of evaluation by combining visual and textual information. Specialized datasets like PolyMath~\cite{polymath} evaluates general cognitive reasoning in multimodal settings, requiring models to combine visual pattern recognition with mathematical logic. MathScape~\cite{mathscape} uses real-world photographs for math questions. GeoEval~\cite{zhang2024geoeval} compiled a comprehensive suite of geometry question. SolidGeo~\cite{solidgeo} focuses on solid/3D geometry, whereas VisAidMath~\cite{visaidmath} evaluates the use of visual aids such as drawing auxiliary lines in geometry. In a different vein, MMSciBench~\cite{ye2025mmscibench} evaluates multimodal scientific problems, covering both math and physics questions in textual and text+image formats. Beyond static images, researchers have explored temporal problem-solving paradigms: VideoMathQA ~\cite{videomathqa} assesses long-horizon mathematical reasoning in educational videos, and NewtonBench ~\cite{newtonbench} challenges LLM-based agents to discover scientific laws through interactive experimentation. MathOPEval~\cite{mathopeval} requires models to perform fine-grained visual operations by generating executable code like constructing, modifying, or annotating a diagram. DynaMath~\cite{dynamath} introduces dynamic variations of visual math problems to test robustness. MaRVL-QA~\cite{marvlqa} uses unlabeled function plots and geometry transformations as inputs that minimizes textual cues. 

Crucially, as researchers identified the limitations of purely result-oriented evaluation, new benchmarks have placed increasing emphasis on the reasoning process itself. MM-MATH~\cite{sun2024mm} exemplifies this shift by augmenting outcome evaluation with process evaluation. We-Math~\cite{wemath} draws inspiration from human problem-solving principles and decomposes problem into sub-problems aligned with knowledge concepts. Math2Visual~\cite{wang2025generating} introduces the task of generating pedagogically useful diagrams for math word problems. A model is given a math problem in text and must produce a diagram or visual illustration that meaningfully represents the problem’s content. Most recently, Geoint-R1~\cite{wei2025geoint} integrates auxiliary element construction with formal verification, and MathCanvas~\cite{shi2025mathcanvas} enabled a unified multimodal model~\cite{bagel} to learn "when and how" to draw as part of its reasoning process. Despite this rich and rapidly evolving landscape, a critical gap persists. The majority of existing evaluations, even those that are process-oriented, ultimately reduce to either checking a final numerical or categorical answer or scoring a model’s generated text against a reference solution. A fundamental challenge remains unaddressed: the lack of a benchmark that holistically evaluates a model's ability to generate a complete, multi-step, and formally verifiable geometric construction from scratch.

\subsection{The Rise of Unified Multimodal Models}

UMMs signifies a pivotal architectural shift, not merely in the unification of modalities, but more fundamentally in the unification of understanding and generation capabilities within a single framework~\cite{zhang2025unified}. This paradigm moves beyond specialized models towards a generalist capable of both interpreting multimodal inputs and generating structured outputs. Recent examples of UMMs include GPT-4o~\cite{hurst2024gpt} and Gemini 2.5~\cite{comanici2025gemini}, as well as powerful open-source alternatives such as Janus series~\cite{janus,chen2025janus,ma2025janusflow}, Qwen-VL~\cite{bai2023qwen,wang2024qwen2}, and OmniBridge~\cite{omnibridge}, are all designed to seamlessly transition between comprehension and synthesis. Other efforts include Apple’s MM1~\cite{mckinzie2024mm1} and Google's Nano Banana (Gemini 2.5 Flash Image)~\cite{google_gemini25_flash_image_2025}. There are also hybrid systems like MM-ReAct~\cite{yang2023mm}, which augment an LLM with external visual tool interfaces rather than a single unified network. Together, these developments mark a shift from modality-specific AI toward end-to-end multimodal intelligence, with UMMs learning unified representations that support cross-modal understanding and generation.

On the one hand, the understanding capabilities of these models have been rigorously validated across a vast landscape of benchmarks. They have achieved exceptional performance on testbeds like MMMU~\cite{yue2024mmmu} for expert-level knowledge, MME-Unify~\cite{xie2025mme} for holistic perception and cognition, and MathVista~\cite{mathvista} for visual mathematical reasoning. Models explicitly designed for deep comprehension, such as Query-Kontext~\cite{querykontext} and MAGUS~\cite{magus}, further demonstrate this proficiency. On the other hand, the generative prowess of UMMs is rapidly advancing, moving far beyond simple image captioning. Models like Bagel~\cite{bagel} and its successor Hyper-Bagel~\cite{hyperbagel} showcase sophisticated generative abilities honed by novel training techniques. Application-specific models like ChartSketcher~\cite{chartsketcher} can synthesize data visualizations from text, while generative models like Lingshu~\cite{lingshu}, VARGPT~\cite{vargpt}, and HaploOmni~\cite{haploomni} exhibit advanced faculties for creating rich, context-aware multimodal content. The development of sophisticated evaluation frameworks like UniEval~\cite{li2025unieval} and reward models such as Skywork-VL Reward~\cite{skyworkvlreward} reflects a growing effort to quantify and guide complex generative behaviors. Despite these parallel advancements in understanding and generation, a critical gap persists in existing evaluation. Benchmarks have largely focused on assessing these two functions as separate capabilities. Consequently, there remains a lack of a framework for evaluating a model's ability to integrate these functions in a \textbf{generative reasoning} task—that is, tasks that \textit{require a model to first deeply comprehend multimodal premises and then generate a structured, logically sound, and verifiable artifact as a solution}. 

\subsection{Code-based Evaluation for Reasoning Tasks}

To enable rigorous evaluation, a growing body of work adopts code as a structured mechanism for assessment, leveraging the verifiable nature of code over free-form natural language explanations. MathCoder-VL~\cite{mathcodervl} employs image-to-code supervision to ensure precise visual–textual alignment, while MATP-BENCH~\cite{matpbench} integrates formal theorem proving in Lean, Coq, and Isabelle for verifiable multimodal logic. VeriEquivBench~\cite{veriequivbench} proposes an equivalence-based validation metric for verifying consistency between generated code and specifications. InternLM-Math~\cite{internlmmath} unifies solving and verifying under a single framework with interleaved reasoning and code execution. DeepMath-103K~\cite{deepmath103k} provides a large, contamination-free dataset with verifiable answers for reinforcement learning with rule-based rewards. MathQ-Verify~\cite{mathqverify} targets problem validity itself through a structured verification pipeline. CMMaTH~\cite{cmmath} introduces a large Chinese multimodal math benchmark and an open-source automatic grader, while MARIO Eval~\cite{marioeval} standardizes automated evaluation across mathematical datasets. U-MATH~\cite{umath} assess university-level reasoning and meta-evaluate LLM judges. In parallel, QuesCo~\cite{quesco} captures holistic mathematical intent through contrastive pretraining.

While recent multimodal datasets~\cite{mathvista,wemath,mathscape,mathv,mathllava,mvmath} have advanced the field, they seldom enforce a verifiable alignment between the natural language reasoning steps. To bridge this gap, GGBench extends the code-based evaluation to the visual geometric domain, enabling a holistic verification pipeline that tightly aligns textual reasoning, code execution, and visual output. This multimodal, verifiable setup allows GGBench to move beyond pattern matching and surface-level correctness.

\section{The GGBench Benchmark}

\subsection{Overview}

We introduce GGBench, a geometric generative reasoning benchmark designed to evaluate UMMs through end-to-end construction tasks. Unlike conventional discriminative benchmarks that evaluate models solely on final answer selection or numerical computation, GGBench operationalizes the assessment of whether a system can genuinely construct a geometrically valid solution.

\begin{figure}[h]
  \centering
  \includegraphics[width=\linewidth]{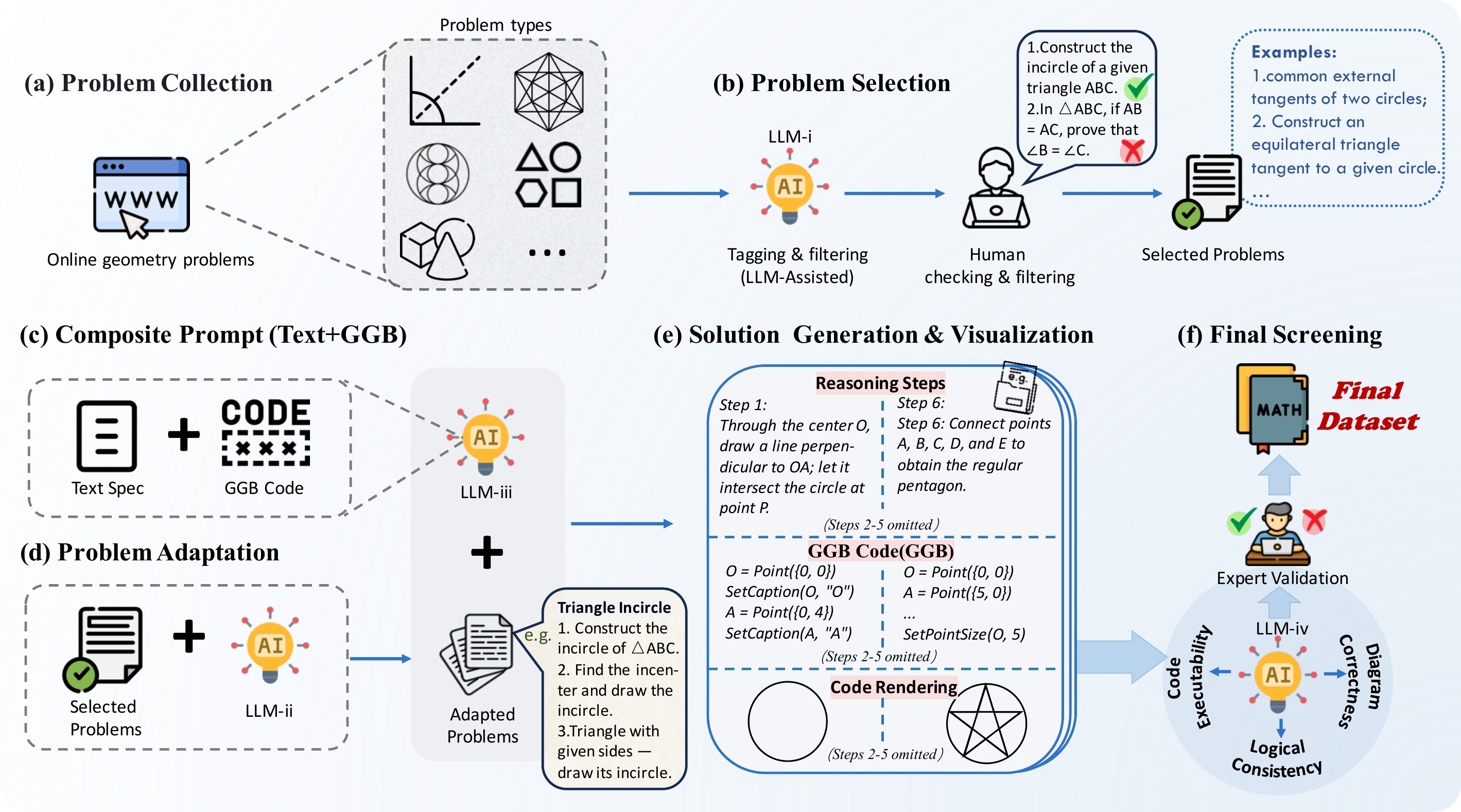}
  \caption{Overview of the GGBench data construction pipeline. }
  \vspace{-4mm}
  \label{fig:pipeline}
\end{figure}

A defining feature of GGBench is its tri-modal data structure, where each sample comprises precisely aligned text, code, and image components. This multi-faceted design enables a uniquely versatile and diagnostic evaluation framework. For instance, the complete, end-to-end task of generating a final image from a textual prompt serves as the primary test for the integrated generative reasoning of UMMs. Concurrently, the benchmark can be decoupled to probe specific capabilities: the natural language reasoning steps (text) and their translation into executable programs (code) can be used to assess the logical planning and code generation faculties of LLMs. Furthermore, the provided GeoGebra code acts as an unambiguous, machine-verifiable ground truth, offering a deterministic method for verifying the geometric correctness and precision of any generated figure. An overview of the multi-modal dataset construction pipeline is shown in Figure~\ref{fig:pipeline}, which defines the stage annotations (a)–(f) referenced throughout the paper.

\subsection{Dataset construction}

\noindent\textbf{Data collection and candidate pooling}
We manually search the web for public geometry problems and assemble a pool that spans classical constructions and contest-style tasks (Figure~\ref{fig:pipeline}, stage~(a)). To scale selection beyond purely manual effort, we use an LLM for assisted tagging and filtering (LLM-i in the figure) and then conduct human checking to retain items that are suitable for construction—i.e., problems with unambiguous geometric dependencies and diagrammatically actionable conditions (Figure~\ref{fig:pipeline}, stage~(b)). This stage emphasizes diversity in objects and relations (e.g., circles, tangents, incenters, parallels) while discarding under-specified prompts.

\noindent\textbf{Prompt design and problem adaptation}
To elicit end-to-end construction in GeoGebra only, we design a \emph{composite} prompt that pairs a textual specification with exemplar GGB code (Figure~\ref{fig:pipeline}, stage~(c)). In parallel, we prepare a reformulation prompt family that converts the selected problems into construction-oriented statements with explicit auxiliary objects and ordered dependencies (Figure~\ref{fig:pipeline}, stage~(d)). In practice, GPT-5~\cite{OpenAI2025GPT5SystemCard} serves as the primary authoring model during adaptation (LLM-ii), while GGB-oriented prompt templates are informed by prior engineering with Gemini~2.5~Pro~\cite{comanici2025gemini} to improve syntactic fidelity. The outcome is a set of adapted problems whose reasoning steps align one-to-one with executable GGB operations.

\noindent\textbf{Solution generation and visualization}
Given an adapted problem and its data-generation prompts, GPT-5~\cite{OpenAI2025GPT5SystemCard} (LLM-iii) produces synchronized outputs: (i) stepwise reasoning text, (ii) executable GGB code, and (iii) a rendered diagram obtained by running the code (Figure~\ref{fig:pipeline}, stage~(e)). This stage focuses on faithful construction: steps materialize auxiliary objects (e.g., angle bisectors, tangents, parallels), code reflects those steps, and the rendered figure visualizes the target configuration. At this point we obtain roughly \( \sim 10{,}000 \) draft instances prior to filtering.

\noindent\textbf{Automated screening and expert finalization}
The final stage involved a rigorous two-tier filtering process to ensure dataset quality. First, we perform LLM-based quality control (LLM-iv) along three axes—\emph{code executability}, \emph{logical consistency of the construction process}, and \emph{diagram correctness with respect to the intended figure}—to remove low-quality drafts (Figure~\ref{fig:pipeline}, stage~(f)). Subsequently, all surviving instances underwent a final review by domain experts who verified their geometric correctness, the sufficiency of the construction, and the precise consistency across the text, code, and image modalities. After this two-tier filtering, GGBench retains \(\,1{,}411\,\) high-quality items with tightly aligned text–code–image triplets. A detailed statistical analysis of the dataset is provided as below.

\begin{table}[ht]
  \centering
  \caption{Corpus-level statistics of GGBench.}
  \label{tab:GGBench_corpus_stats}
  \setlength{\tabcolsep}{12mm}{
  \begin{tabular}{lrr}
    \toprule
    \multicolumn{3}{l}{\textbf{Dataset size}} \\
    \midrule
    Total problems & \multicolumn{2}{r}{1{,}411} \\
    \midrule
    \multicolumn{3}{l}{\textbf{Problem types}} \\
    Straightedge-and-compass construction & \multicolumn{2}{r}{798} \\
    Geometric transformation construction & \multicolumn{2}{r}{426} \\
    Analytic construction & \multicolumn{2}{r}{187} \\
    \midrule
    \multicolumn{3}{l}{\textbf{Difficulty levels}} \\
    Easy  & \multicolumn{2}{r}{298} \\
    Medium & \multicolumn{2}{r}{816} \\
    Hard  & \multicolumn{2}{r}{297} \\
    \midrule
    \multicolumn{3}{l}{\textbf{Images per problem}} \\
    3 images & \multicolumn{2}{r}{7} \\
    4 images & \multicolumn{2}{r}{227} \\
    5 images & \multicolumn{2}{r}{860} \\
    6 images & \multicolumn{2}{r}{283} \\
    7 images & \multicolumn{2}{r}{34} \\
    Total images & \multicolumn{2}{r}{7{,}165} \\
    \midrule
    \multicolumn{3}{l}{\textbf{Question length (tokens)}} \\
    Minimum & \multicolumn{2}{r}{68} \\
    Maximum & \multicolumn{2}{r}{483} \\
    Average & \multicolumn{2}{r}{189.83} \\
    \bottomrule
  \end{tabular}
  }
\end{table}

\subsection{Dataset Analysis}

\paragraph{Dataset Composition and Scale}
GGBench comprises 1{,}411 GeoGebra-based construction problems that integrate reasoning, code, and visual outputs. Table~\ref{tab:GGBench_corpus_stats} summarizes overall dataset statistics, including problem types, difficulty levels, image counts, and question lengths. Most problems belong to the medium difficulty (57.83\%), balancing solvability and reasoning complexity. Straightedge-and-compass constructions dominate (56.6\%), followed by geometric transformation and analytic constructions. Each problem includes between 3 and 7 diagrams, averaging 5.08 images per problem, providing dense visual evidence for evaluating generative reasoning.

\begin{table}[ht]
  \centering
  \caption{Distribution of reasoning categories in GGBench. Each problem may involve multiple categories, yielding 3,097 total tags across 1,411 problems.}
  \label{tab:GGBench_categories}
  \setlength{\tabcolsep}{12mm}
  \begin{tabular}{lr}
    \toprule
    \textbf{Category} & \textbf{Count} \\
    \midrule
    Basic Constructions & 1{,}063 \\
    Circle Properties \& Constructions & 931 \\
    Geometric Transformations & 376 \\
    Triangle Properties \& Constructions & 280 \\
    Applications of Geometric Theorems & 218 \\
    Polygon Properties \& Constructions & 107 \\
    Measurement \& Ratios & 92 \\
    Locus Constructions & 30 \\
    \midrule
    Total category tags & 3{,}097 \\
    \bottomrule
  \end{tabular}
\end{table}
\begin{figure}[t]
  \centering
  \includegraphics[width=\linewidth]{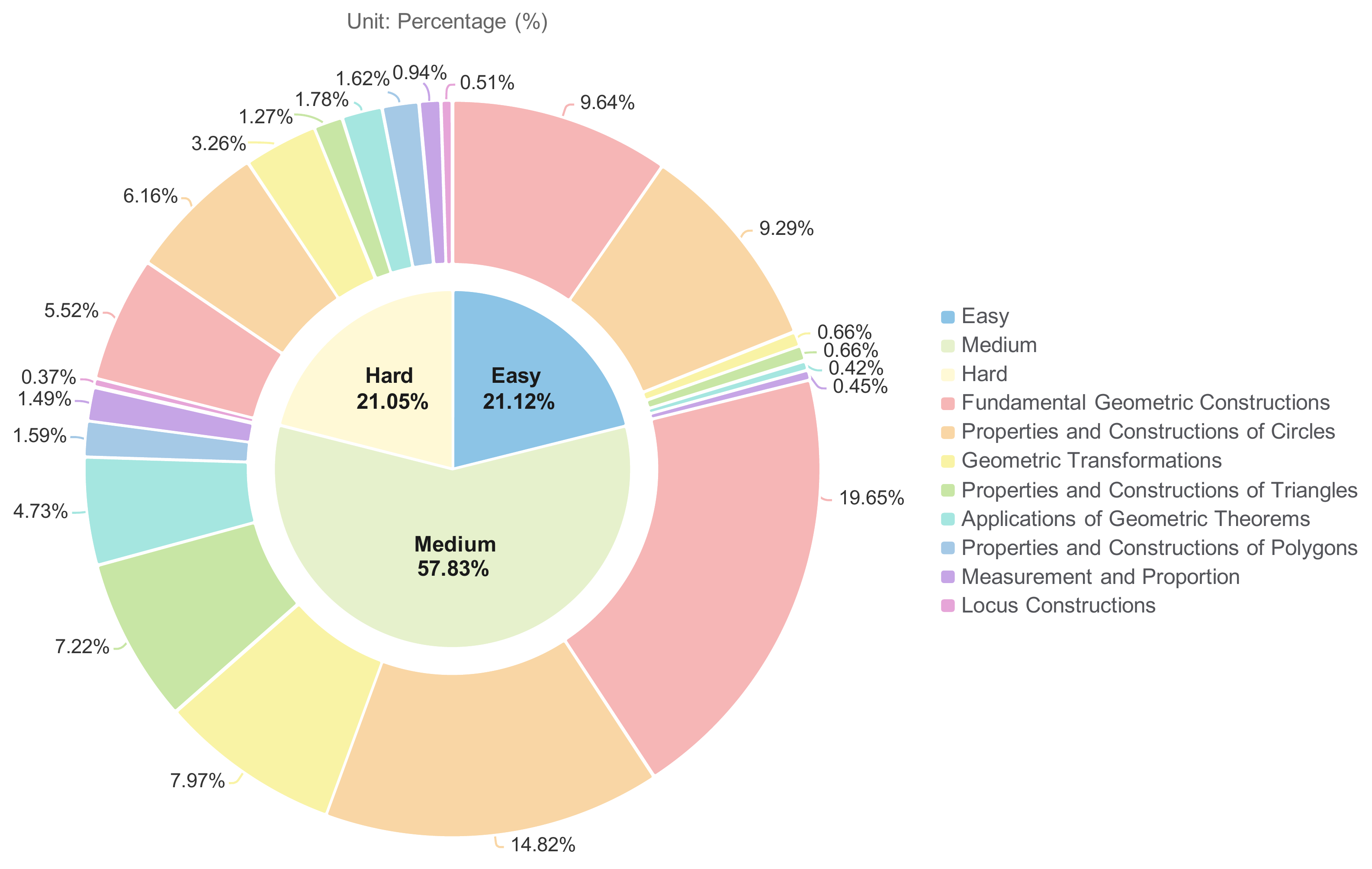}
  \caption{Difficulty distribution and category composition in GGBench. The inner ring shows the proportion of difficulty levels (Easy/Medium/Hard), while the outer ring presents category shares within each difficulty band, reflecting progressive complexity across reasoning types.}
  \label{fig:difficulty_breakdown}
\end{figure}

\paragraph{Category Distribution and Difficulty}
GGBench covers eight major categories of geometric reasoning, as detailed in Table~\ref{tab:GGBench_categories}. Each problem may involve multiple skills, resulting in 3{,}097 category annotations across the dataset. As shown in Figure~\ref{fig:difficulty_breakdown}, \emph{Basic Constructions} and \emph{Circle Properties \& Constructions} are predominant across all difficulty levels, while advanced reasoning types—such as \emph{Applications of Geometric Theorems} and \emph{Measurement \& Ratios}—become more frequent in the hard subset. This stratification ensures balanced coverage from elementary to advanced geometric reasoning.

\paragraph{Category Composition and Cognitive Coverage}
The distribution in Table~\ref{tab:GGBench_categories} reflects a deliberate cognitive gradient. The majority of tasks emphasize procedural reasoning (e.g., bisectors, parallels, and tangents), providing a stable foundation for evaluating geometric comprehension. Intermediate categories, including \emph{Transformations} and \emph{Triangles}, introduce abstraction through relational constraints and inter-object dependencies. Higher-order categories—such as \emph{Theorem Applications} and \emph{Measurement \& Ratios}—demand symbolic reasoning, multi-step dependencies, and precision alignment between text and geometry. This stratification enables a fine-grained evaluation of a model's capabilities, from foundational geometric execution to advanced, abstract problem-solving.

\paragraph{Token Length and Reasoning Path}
We report question length in tokens. As shown in Table~\ref{tab:GGBench_corpus_stats}, question length ranges from 68 to 483 tokens, averaging 189.83 tokens. This range reflects the multi-step reasoning nature of GGBench tasks, requiring the model to plan, describe, and validate geometric constructions.
On average, each problem contains 5.08 rendered images (7{,}165/1{,}411), and carries 2.20 category tags (3{,}097/1{,}411), indicating that GGBench is both visually dense and multi-skill by design. The five-image setting is the most common (60.9\%; 860/1{,}411), providing ample visual evidence for stepwise construction. While \emph{Basic Constructions} and \emph{Circle Properties \& Constructions} dominate across the corpus, Figure~\ref{fig:difficulty_breakdown} further shows a progressive shift toward higher-order skills (e.g., \emph{Applications of Geometric Theorems}, \emph{Measurement \& Ratios}) in the hard subset.

\subsection{Comparison with Existing Benchmarks}
\label{subsec:GGBench_comparison}

Table~\ref{tab:benchmark_comparison} situates GGBench among representative multimodal math benchmarks along two groups of axes: (i) \emph{capability coverage}---the proportion of samples that exercise \emph{understanding}, \emph{generation}, and \emph{multi-step} reasoning; and (ii) \emph{modality supervision}---whether problems provide aligned \emph{text}, \emph{images}, and \emph{code}. We normalize modality coverage as a binary indicator in the table (\ding{51}/\ding{55}) for readability.

\begin{table}[t]
  \centering
  \caption{Comparison with existing multimodal mathematical benchmarks. Percentages under \emph{Understanding/Generation/Multi-step} indicate the share of samples exercising each capability. \emph{Text/Image/Code} report modality support (\%). GGBench uniquely achieves full tri-modal coverage with executable code.}
  \label{tab:benchmark_comparison}
  \setlength{\tabcolsep}{2.8mm}{
  \begin{tabular}{lcccccc}
    \toprule
    \textbf{Benchmark} & \textbf{Understanding} & \textbf{Generation} & \textbf{Multi-step} & \textbf{Text} & \textbf{Image} & \textbf{Code} \\
    \midrule
    MathVista~\cite{mathvista}      & 49.9 & 34.8 & 21.5 & \ding{51}
 & \ding{51} & \ding{55} \\
    MathVerse~\cite{mathverse}      & 61.3 & 52.7 & 45.2 & \ding{51} & \ding{51} & \ding{55} \\
    MM-MATH~\cite{sun2024mm}        & 31.0 & 26.5 & 82.0 & \ding{51} & \ding{51} & \ding{55} \\
    MathScape~\cite{mathscape}      & 44.2 & 39.8 & 18.5 & \ding{51} & \ding{51} & \ding{55} \\
    GeoEval~\cite{zhang2024geoeval}        & 46.0 & 25.0 & 33.3 & \ding{51} & \ding{51} & \ding{55} \\
    WE-MATH~\cite{wemath}        & 58.7 & 51.9 & 42.6 & \ding{51} & \ding{51} & \ding{55} \\
    MMSciBench~\cite{ye2025mmscibench}     & 55.4 & 32.8 & 40.2 &  \ding{51} & \ding{51} & \ding{55} \\
    MATH2VISUAL~\cite{wang2025generating}     & 48.0 & 37.0 & \ding{55} & \ding{51} & \ding{51} & \ding{55} \\
    PolyMath~\cite{polymath}       & 54.6 & \ding{55} & \ding{55} & \ding{51} & \ding{55} & \ding{55} \\
    SOLIDGEO~\cite{solidgeo}       & 49.5 & 18.7 &  6.7 & \ding{51} & \ding{51} & \ding{55} \\
    \midrule
    GGBench       & 100.0  & 100.0  & 100.0  & \ding{51} & \ding{51} & \ding{51} \\
    \bottomrule
  \end{tabular}
  }
\end{table}

\paragraph{Generative construction vs.\ answer selection}
Most benchmarks~\cite{mathvista,mathscape} emphasize recognition or answer selection. Even when generation is present~\cite{mathverse}, outputs are rarely required to be \emph{constructive artifacts} that satisfy constraints. By contrast, every GGBench instance requires producing diagrams with formal constraints, shifting evaluation from choosing an answer to \emph{constructing evidence}.

\paragraph{Tri-modal alignment and verifiability}
A central gap in prior work is the absence of code supervision. It is difficult to verify whether a model’s reasoning aligns with the final diagram. GGBench provides 100\% alignment between \emph{text} (stepwise plan), \emph{code} (executable construction), and \emph{image} (rendered diagram), enabling process-level judging, program executability checks, and geometry-aware verification within a single benchmark.

\paragraph{Multi-step supervision}
While some datasets include multi-step elements~\cite{sun2024mm,mathverse}, they do not couple each step with executable constructions, limiting verifiability. GGBench enforces multi-step reasoning at \emph{100\%} and ties steps to code, ensuring that auxiliary objects and dependencies are operationalized rather than narrated.

\paragraph{Modal breadth and heterogeneity.}
Most benchmarks offer full text and image coverage (100\%/100\%), with exceptions such as MMSciBench~\cite{ye2025mmscibench} (90\%/40\%) and PolyMath~\cite{polymath} (100\%/0\%). GGBench maintains complete modality support (text/image/code all 100\%), so reasoning, rendering, and execution can be evaluated cohesively.

Together, these dimensions position GGBench as a construction-centric benchmark that unifies understanding, reasoning, and generation under a verifiable framework. It complements prior understanding-oriented datasets, offering a rigorous testbed for the next generation of unified multimodal reasoning models.

\section{Experiment}
\label{section:experiment}

\paragraph{Setup.}
We evaluate existing Unified Multimodal Models on GGBench to assess their generative geometric reasoning capabilities. Each model receives the problem text and reference image as input. To ensure deterministic outputs, we fix the temperature at 0.0 during inference. This eliminates stochasticity and enforces consistent step-by-step reasoning. All code generation adheres strictly to the GeoGebra command syntax to ensure executability and structural correctness. Inference is parallelized for efficiency. All models are evaluated under identical settings with unified prompts and processing pipelines. Full prompt templates are detailed in Appendix.

\subsection{Baselines}

We evaluate models under two complementary tracks that correspond to distinct architectural paradigms: (A) end-to-end UMMs that directly generate diagram images from natural-language prompts, and (B) LLMs/LRMs that first produce explicit, executable construction code which is then rendered into diagrams. This dual-track evaluation allows us to quantify and analyze the gap between UMMs’ immediate visual generation and the geometrically grounded, code-driven constructions produced by LLMs/LRMs. 

\noindent\textbf{Track A: End-to-end UMMs}
These systems take natural-language specifications and output diagram images step-by-step.
We include: \textit{Qwen-Image}~\cite{wu2025qwen}, \textit{Seedream 4.0}~\cite{seedream2025seedream}, \textit{Janus}~\cite{janus}, \textit{BAGEL}~\cite{bagel}, and \textit{Nano Banana}~\cite{google_gemini25_flash_image_2025}. Among them, \textit{Qwen-Image} and \textit{Seedream~4.0} produce only final images without stepwise explanations, whereas the other models generate multi-stage visual outputs for step evaluation.

\noindent\textbf{Track B: Planning $\rightarrow$ Code $\rightarrow$ Render}
These models first produce a \emph{textual plan} and then emit \emph{GeoGebra code} to render into images: \textit{GPT-4o}~\cite{hurst2024gpt}, \textit{GLM-4.5V}~\cite{hong2025glm}, \textit{Qwen3VL-235B-A22B}~\cite{yang2025qwen3}, \textit{GPT-4}~\cite{achiam2023gpt}, \textit{GPT-5}~\cite{OpenAI2025GPT5SystemCard}, \textit{Claude Sonnet 4.5}~\cite{Anthropic2025ClaudeSonnet4_5}, \textit{Gemini-2.5-Pro}~\cite{comanici2025gemini}, \textit{DeepSeek-R1}~\cite{liu2024deepseek}, \textit{DeepSeek-V3.1}~\cite{liu2024deepseek}, and \textit{Qwen3-14B}~\cite{yang2025qwen3}.

\subsection{Metrics}

To evaluate geometric generative reasoning, We adopt a four-stage protocol: (1) \textit{Planning}, (2) \textit{Middle Process}, (3) \textit{Final Result}, and (4) \textit{Overall Scores}. All automatic scores are produced by a VLM GPT-4o using fixed prompts.

\noindent\textbf{(1) Planning (VLM-T)} measures the model’s ability to formulate a coherent step-by-step plan in natural language before any code or drawing is produced. Given the textual reasoning output, the VLM judge scores each response along three dimensions: (i) Logical coherence, (ii) Step completeness, and (iii) Geometric correctness. Each criterion is rated on a 1–5 scale and rescaled to [0,100]. Details of the judging rubric and prompt design appear in Appendix~\ref{app:vlm_t}.

\noindent\textbf{(2) Middle process (VLM-I-Mid)} To evaluate intermediate reasoning consistency, we concatenate all intermediate construction images into a chronological panel and feed it to the same VLM judge. The model assesses: (i) Step accuracy and (ii) Process consistency. Implementation and prompt details are provided in Appendix~\ref{app:vlm_mid}.

\noindent\textbf{(3) Final result (VLM-I-Res)} assesses the geometric correctness of the final diagram relative to the reference solution. In addition, we also report pixel-level metrics. Details are in Appendix~\ref{app:vlm_res}.

\noindent\textbf{(4) Overall scores} The overall VLM score (VLM–I) is computed as the mean of the intermediate VLM-I-Mid and final scores VLM-I-Res. Human raters follow the same rubric as the VLM judge; inter-rater consistency and calibration procedures are detailed in Appendix~\ref{app:human_eval}. We observe a strong Pearson correlation ($r=0.9295$) between VLM and human scores, validating the high reliability of our automated evaluation for this task.


\subsection{Main Results}

Table~\ref{tab:main_results} summarizes the overall performance across both tracks. End-to-end UMMs remain significantly behind code-driven models in nearly all evaluation dimensions. Even the strongest UMM, \textit{Nano Banana}, ranks only in the middle tier of code-based systems, illustrating that direct visual generation still struggles to enforce geometric constraints. C''ode-level metrics of LLMs/LRMs are reported in Table~\ref{tab:code_results}.

\noindent\textbf{The planning stage (\textbf{VLM-T})} evaluates a model’s capacity to reason step-by-step before any code or drawing is produced. Interestingly, \textit{Nano Banana} achieves planning scores comparable to several LLMs/LRMs. However, models such as \textit{GPT-4o} and \textit{GLM-4.5V}, despite generating coherent plans, often fail to produce executable code, limiting their downstream visual accuracy. This gap demonstrates that high-level reasoning alone is insufficient without grounding in executable geometry.

\noindent\textbf{Intermediate and final constructions (VLM–I)}
Across both the intermediate (\texttt{VLM–I\textsubscript{mid}}) and final (\texttt{VLM–I\textsubscript{res}}) stages, code-driven models consistently outperform end-to-end generators. 
Explicitly reasoning through textual plans and producing executable constructions leads to markedly higher geometric correctness and visual coherence. 
While robust planning does not always guarantee flawless drawings, weak or inconsistent reasoning almost always results in invalid geometry—reinforcing the necessity of tightly coupled reasoning–generation pipelines.
Moreover, purely pixel-based metrics (PSNR, SSIM, LPIPS) show only weak correlation with geometric validity: 
high perceptual similarity can mask structural errors such as misaligned intersections or missing tangencies.
Hence, evaluation based on verifiable geometry, rather than appearance alone, offers a more reliable measure of multimodal reasoning quality.

\begin{table*}[t]
\centering
\small
\caption{
Main results on GGBench. Higher is better~(↑) except for LPIPS~(↓).
}
\label{tab:main_results}
\setlength{\tabcolsep}{0.8mm}{
\begin{tabular}{l|c|c|cccc|cc}
\toprule
\multirow{2}{*}{\textbf{Model}} 
& \multicolumn{1}{c|}{\textbf{Planning}} 
& \multicolumn{1}{c|}{\textbf{Middle Process}} 
& \multicolumn{4}{c|}{\textbf{Final Result}} 
& \multicolumn{2}{c}{\textbf{Overall Scores}} \\
\cmidrule(lr){2-2}\cmidrule(lr){3-3}\cmidrule(lr){4-7}\cmidrule(lr){8-9}
& {VLM-T~↑} & {VLM-I-Mid~↑} & {VLM-I-Res~↑} & {LPIPS~↓} & {PSNR~↑} & {SSIM~↑} & {VLM-I~↑} & {Human~↑} \\
\midrule
\rowcolor[HTML]{EFEFEF}
\multicolumn{9}{c}{\textit{End-to-end UMMs}} \\
Qwen-Image~\cite{wu2025qwen} & - & - & 22.75 & 56.39 & 58.23 & 48.06 & 22.75 & 25.56 \\
Seedream 4.0~\cite{seedream2025seedream} & - & - & 24.45 & 51.06 & 59.44 & 56.44 & 24.45 & 37.56 \\
Janus~\cite{janus} & 33.85 & 21.69 & 19.76 & 57.74 & 57.76 & 60.97 & 20.73 & 19.46 \\
BAGEL~\cite{bagel} & 23.07 & 21.84 & 19.99 & 57.07 & 61.78 & 58.82 & 20.91 & 20.12 \\
Nano Banana~\cite{comanici2025gemini} & 58.54 & 44.83 & 22.81 & 51.85 & 64.53 & 59.51 & \textbf{33.82} & \textbf{45.75} \\
\midrule
\rowcolor[HTML]{EFEFEF}
\multicolumn{9}{c}{\textit{LLMs/LRMs}} \\
GPT-4o~\cite{hurst2024gpt} & 59.73 & 26.19 & 2.66 & 95.43 & 5.45 & 5.69 & 14.43 & 23.04 \\
GLM-4.5V~\cite{hong2025glm} & 53.32 & 25.63 & 5.02 & 52.91 & 12.19 & 12.94 & 15.33 & 30.14 \\
Qwen3-14B~\cite{yang2025qwen3} & 58.65 & 39.30 & 12.97 & 78.81 & 23.92 & 24.81 & 26.13 & 38.23 \\
Gemini 2.5 Pro~\cite{comanici2025gemini} & 38.50 & 37.41 & 15.80 & 68.39 & 37.17 & 39.73 & 26.61 & 44.68 \\
DeepSeek-R1~\cite{liu2024deepseek} & 61.16 & 62.42 & 20.48 & 66.06 & 37.94 & 37.59 & 41.45 & 49.55 \\
GPT-4~\cite{achiam2023gpt} & 55.66 & 50.99 & 15.10 & 67.35 & 35.26 & 38.31 & 33.04 & 55.99 \\
Qwen3-VL~\cite{yang2025qwen3} & 56.40 & 49.55 & 23.94 & 39.40 & 52.33 & 58.71 & 36.74 & 66.77 \\
DeepSeek-V3.1~\cite{liu2024deepseek} & 60.24 & 73.13 & 26.41 & 57.21 & 48.33 & 50.12 & 49.77 & 68.12 \\
Claude Sonnet 4.5~\cite{Anthropic2025ClaudeSonnet4_5} & 61.19 & 77.92 & 30.29 & 52.22 & 51.74 & 50.52 & 54.11 & 72.12 \\
GPT-5~\cite{OpenAI2025GPT5SystemCard} & 62.01 & 76.79 & 37.36 & 49.65 & 54.80 & 59.49 & \textbf{57.08} & \textbf{83.06} \\
\bottomrule
\end{tabular}}
\end{table*}

\begin{table*}[t]
\centering
\small
\caption{Evaluation results on GGBench-Code across execution, similarity, and structural metrics.}
\label{tab:code_results}
\setlength{\tabcolsep}{0.9mm}{
\resizebox{\linewidth}{!}{
\begin{tabular}{l|c|cccc|c}
\toprule
\textbf{Model} & \textbf{Pass@1~↑} & \textbf{BLEU Mean~↑} & \textbf{RUBY Mean~↑} & \textbf{ROUGE-L Mean~↑} & \textbf{chrF Mean~↑} & \textbf{EditDist Mean~↓} \\
\midrule
GPT-4o~\cite{hurst2024gpt} & 7.87 & 8.33 & 12.85 & 27.41 & 29.17 & 78.92 \\
GLM-4.5V~\cite{hong2025glm} & 14.25 & 10.28 & 17.53 & 30.48 & 38.32 & 79.61 \\
Qwen3-14B~\cite{yang2025qwen3} & 34.30 & 14.04 & 23.85 & 38.07 & 44.36 & 76.89 \\
Gemini 2.5 Pro~\cite{comanici2025gemini} & 32.67 & 24.21 & 40.56 & 50.50 & 65.87 & 71.14 \\
DeepSeek-R1~\cite{liu2024deepseek} & 54.36 & 18.46 & 33.68 & 42.69 & 60.11 & 73.64 \\
GPT-4~\cite{achiam2023gpt} & 50.53 & 17.31 & 28.54 & 44.62 & 51.25 & 70.34 \\
Qwen3-VL-~\cite{yang2025qwen3} & 69.31 & 23.24 & 38.61 & 50.73 & 63.51 & 70.24 \\
DeepSeekV3.1~\cite{liu2024deepseek} & 70.09 & 20.44 & 35.14 & 47.36 & 59.24 & 70.87 \\
Claude Sonnet 4.5~\cite{Anthropic2025ClaudeSonnet4_5} & 75.34 & 22.04 & 36.77 & 49.68 & 60.28 & 70.81 \\
GPT-5~\cite{OpenAI2025GPT5SystemCard} & 79.02 & 18.87 & 39.92 & 44.08 & 62.44 & 74.13 \\
\bottomrule
\end{tabular}}}
\end{table*}

\noindent\textbf{Code-based analysis (Table~\ref{tab:code_results})}
At the program level, \textit{GPT-5} achieves the highest execution accuracy (\texttt{pass@1}=79.02), followed closely by \textit{Claude Sonnet~4.5} (75.34) and \textit{DeepSeek-V3.1} (70.09). 
Interestingly, code-similarity metrics (e.g., BLEU or ROUGE-L) do not necessarily predict executable success: \textit{Gemini-2.5-Pro} attains high textual similarity yet a lower \texttt{Pass@1}, confirming that surface-level resemblance does not imply geometric equivalence.
This observation validates the necessity of execution-based evaluation in GGBench.

\noindent\textbf{Summary of findings}
In summary, reasoning-grounded models that produce and execute code achieve substantially higher geometric correctness, interpretability, and human-rated quality.
\textit{GPT-5} attains the best overall performance (\texttt{VLM–I}=57.08, \texttt{Human}=83.06), followed by \textit{Claude Sonnet~4.5} and \textit{DeepSeek-V3.1}. 
Despite recent advances, purely generative UMMs remain limited in enforcing spatial constraints, emphasizing the importance of explicit, verifiable construction pipelines for next-generation multimodal reasoning systems.

\subsection{Performance by Construction Category}\

\begin{figure}[ht]
    \centering
    \includegraphics[width=\linewidth]{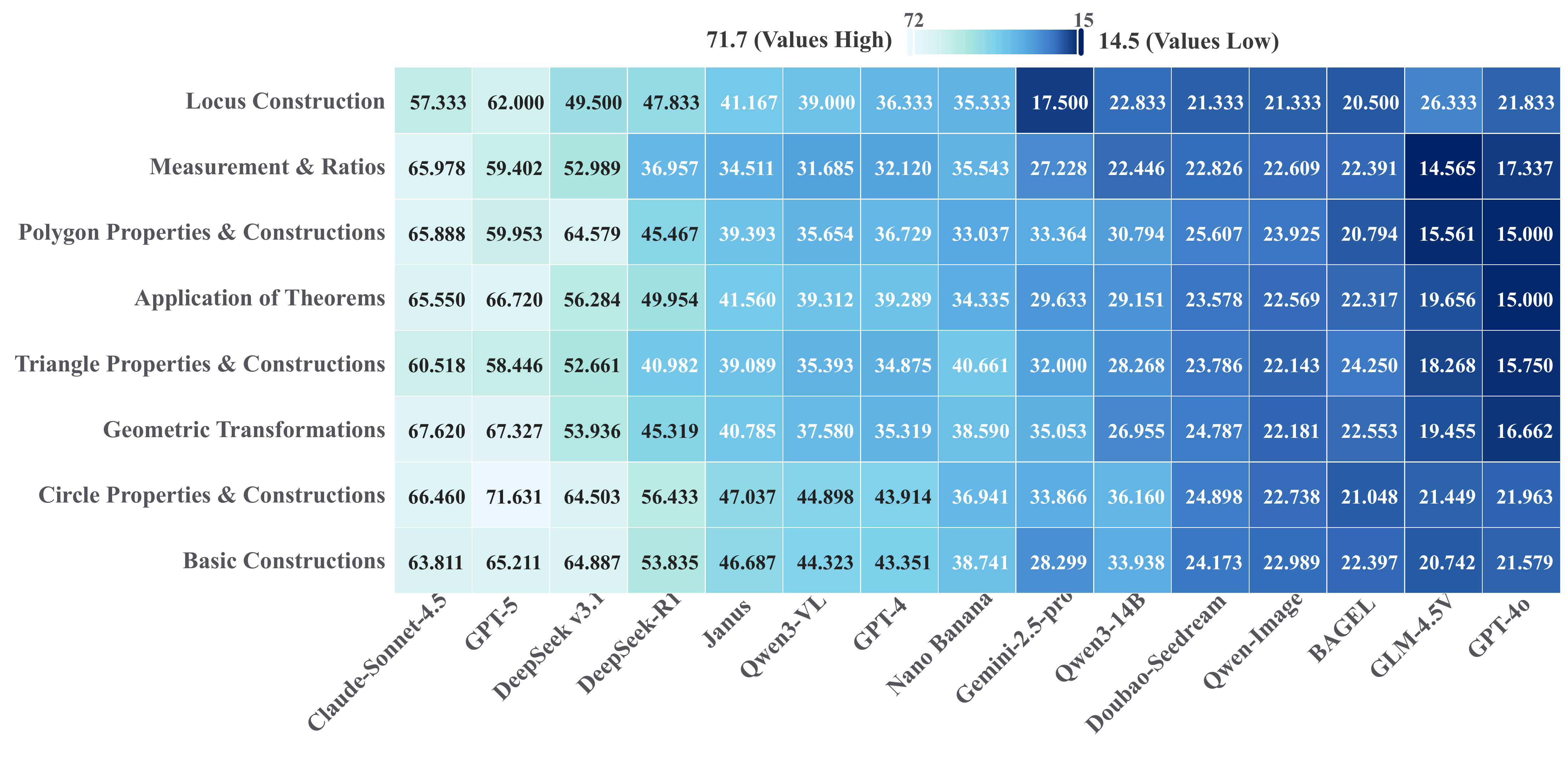}
    \caption{VLM-I scores across eight construction categories in GGBench. Each cell reflects the average multimodal reasoning quality for a model-category pair.}
    \label{fig:category_heatmap}
\end{figure}

\autoref{fig:category_heatmap} reports the average VLM–I scores across eight construction categories in GGBench, revealing distinct performance patterns among models.
{GPT-5}~\cite{OpenAI2025GPT5SystemCard} achieves the highest overall accuracy, consistently outperforming other models across all categories. 
Its strength lies in maintaining both \emph{symbolic precision} (accurate translation from plan to executable code) and \emph{spatial coordination} (faithful geometric realization).
{Claude-Sonnet-4.5}~\cite{Anthropic2025ClaudeSonnet4_5} ranks second, closely followed by {DeepSeek-V3.1}~\cite{liu2024deepseek}, particularly excelling in tasks requiring geometric regularity such as \emph{Geometric Transformations} and \emph{Circle Properties \& Constructions}. 
These models exhibit stable reasoning across categories, indicating robust integration between linguistic and spatial modalities. {Qwen3-VL}~\cite{yang2025qwen3} demonstrates strong performance on structurally constrained categories—such as \emph{Basic Constructions} and \emph{Triangle Properties \& Constructions}—where relationships among objects are more rule-based, but its scores decline in \emph{Measurement \& Ratios} and \emph{Applications of Theorems}, which require quantitative and theorem-level reasoning. 
This contrast suggests that while Qwen3-VL effectively grounds syntactic geometry, it struggles with deeper symbolic arithmetic integration. At the lower end, vision-first models such as {GPT-4o}~\cite{hurst2024gpt}, {GLM-4.5V}~\cite{hong2025glm}, and {Qwen-Image}~\cite{wu2025qwen} consistently underperform, with mean scores below~23 across most categories.
Their weakness is especially evident in tasks requiring explicit relational constraints or quantitative precision, implying insufficient geometric grounding despite competent visual synthesis. Across all models, the most challenging categories are \emph{Measurement \& Ratios} and \emph{Applications of Theorems}, where performance drops by 10–15 points relative to simpler procedural tasks.
These categories require not only visual generation but also algebraic and deductive reasoning, emphasizing the need for symbolic–geometric alignment.
Conversely, \emph{Basic Constructions} and \emph{Circle Properties} yield the highest scores overall, indicating that most models can handle canonical constructions but fail to generalize to theorem-driven or quantitative reasoning.

\paragraph{Discussion}
The category-wise trends highlight the diagnostic value of GGBench.
By decomposing performance along interpretable geometric skills, the benchmark exposes which reasoning components—procedural, transformational, or quantitative—limit current multimodal systems.
The results confirm that achieving high-fidelity diagram generation alone does not guarantee correct geometric reasoning, underscoring the necessity of construction-level, verifiable evaluation frameworks.

\subsection{Performance by Question Type}

\begin{figure}[h]
    \centering
    \includegraphics[width=0.6\linewidth]{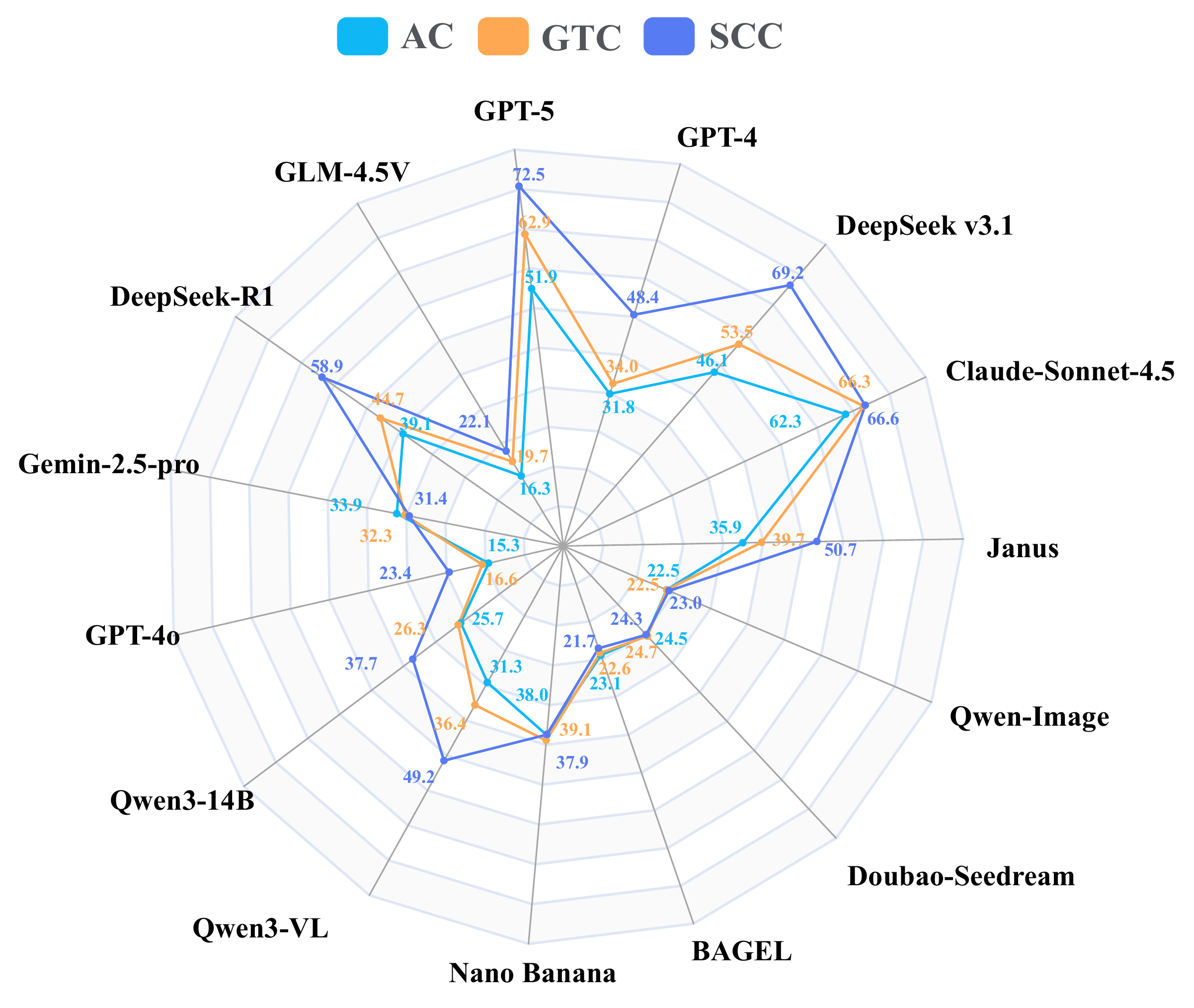}
    \caption{VLM-I scores across geometric task types: Analytic Construction (AC), Geometric Transformation Construction (GTC), and Straightedge-and-Compass Construction (SCC). Higher values indicate better performance.}
    \label{fig:question_type_radar}
\end{figure}

\autoref{fig:question_type_radar} presents the performance of all evaluated models across three geometric task types: analytic construction (AC), geometric transformation construction (GTC), and straightedge-and-compass construction (SCC), measured using the unified VLM-I metric. Clear distinctions emerge across categories. Models generally perform best on SCC tasks that emphasize rule-based geometric procedures, where GPT-5~\cite{OpenAI2025GPT5SystemCard} reaches 72.54, followed closely by Claude-Sonnet-4.5~\cite{Anthropic2025ClaudeSonnet4_5} and DeepSeek V3.1~\cite{liu2024deepseek}, both above 66. Performance on GTC tasks is moderately lower: Claude-Sonnet-4.5 records 66.26 and DeepSeek V3.1 achieves 53.53, reflecting the increased complexity of maintaining geometric invariants under rotation or reflection. Analytic construction tasks yield the lowest results, with GPT-5 scoring 51.90 and DeepSeek V3.1 reaching 46.07, likely due to the greater spatial flexibility and reduced structural constraints of such problems. UMMs such as Janus~\cite{janus}, BAGEL~\cite{bagel}, and Seedream~\cite{seedream2025seedream} remain consistently weaker due to the absence of symbolic reasoning, while text-only systems like DeepSeek V3.1 show competitive performance on structure-driven categories through accurate code synthesis. Overall, these findings indicate that model effectiveness is shaped jointly by modality alignment and the specific geometric reasoning paradigm required by the task.

\paragraph{Discussion}
Overall, performance varies systematically with the degree of symbolic structure embedded in the task.
Tasks governed by explicit geometric rules (SCC) favor models with strong language-to-code translation ability, whereas transformation and analytic tasks expose weaknesses in maintaining geometric invariants and algebraic reasoning.
These results underscore that multimodal effectiveness in GGBench is shaped jointly by a model’s modality alignment and by the specific reasoning paradigm—procedural, transformational, or analytic—required by the construction type.

\subsection{Performance by Difficulty Level}
\begin{figure}[t]
    \centering
    \includegraphics[width=0.96\linewidth]{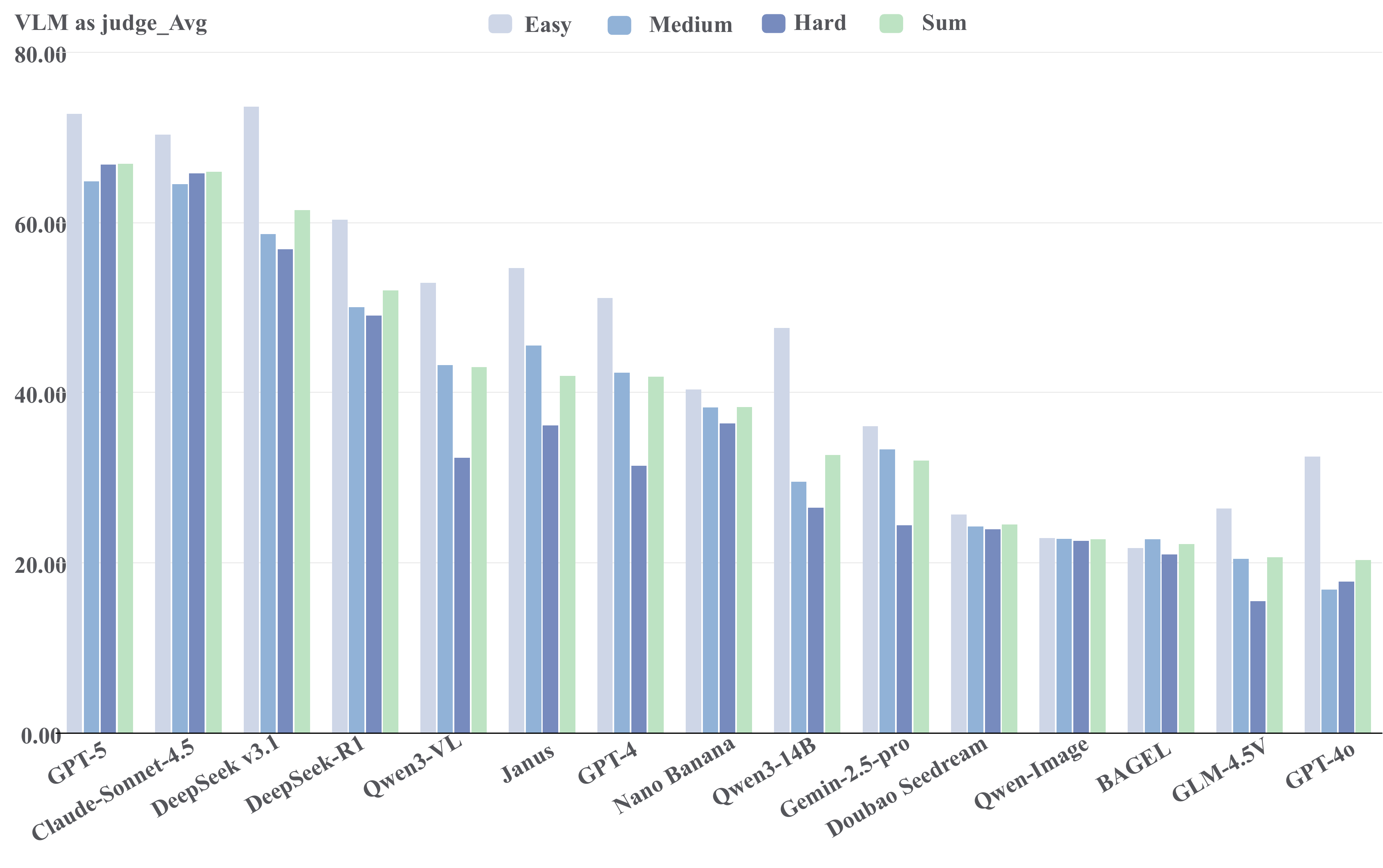}
    \caption{VLM-I performance across difficulty levels on GGBench. Bars represent \textit{Easy}, \textit{Medium}, \textit{Hard}, and overall scores. VLM-I captures both intermediate reasoning quality and final visual correctness.}
    \label{fig:difficulty_lpips}
\end{figure}

\autoref{fig:difficulty_lpips} presents the VLM-I scores across three difficulty levels—Easy, Medium, and Hard—providing a comprehensive view of how reasoning complexity impacts model performance. The majority of systems perform best on Easy problems and experience a steady decline as task difficulty increases, confirming the progressive design of GGBench. This pattern is especially pronounced for models like Qwen3-14B~\cite{yang2025qwen3} and GPT-4~\cite{achiam2023gpt}, which drop more than 20 points from Easy to Hard, suggesting that their performance is disproportionately sensitive to longer reasoning chains and higher geometric abstraction. In contrast, GPT-5~\cite{OpenAI2025GPT5SystemCard} maintains strong results across all difficulty tiers, scoring 72.70 on Easy and 66.77 on Hard, highlighting its robustness in both planning and execution. Claude-Sonnet-4.5~\cite{Anthropic2025ClaudeSonnet4_5} follows a similar trend, while DeepSeek V3.1~\cite{liu2024deepseek} notably surpasses both on Easy problems with a peak of 73.57. UMMs like Nano Banana~\cite{comanici2025gemini}, Janus~\cite{janus}, and Seedream~\cite{seedream2025seedream} show more stable but consistently lower performance across difficulties, indicating weaker adaptation to reasoning complexity despite visual fluency. These observations reinforce that difficulty scaling in GGBench effectively distinguishes between symbolic robustness and shallow pattern matching, making VLM-I a reliable diagnostic signal for geometric reasoning across complexity levels.

\section{Error Analysis}
\begin{figure}[t]
  \centering
  \includegraphics[width=\linewidth]{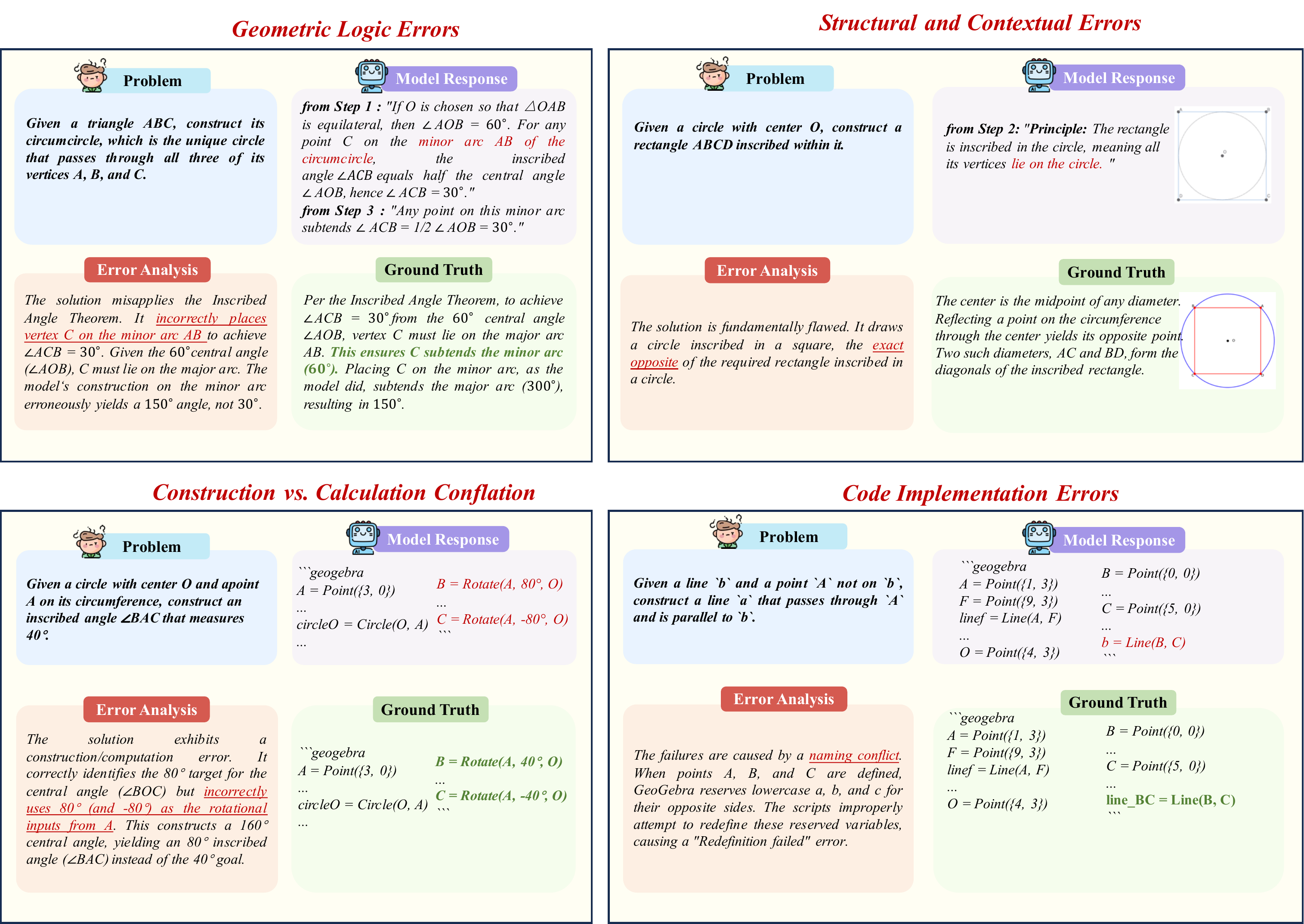}
  \caption{The common error analysis.}
  \label{fig:error_analysis}
\end{figure}

To better understand failure patterns in geometric reasoning, we analyze incorrect outputs from benchmark evaluation and identify four common error types: (i) geometric logic errors, due to misapplied theorems; (ii) structural and contextual errors, involving confusion in figure containment or spatial relationships; (iii) conflation of construction and numeric goals; and (iv) code-level failures from syntax misuse or reserved keyword conflicts.
\autoref{fig:error_analysis} presents a structural and contextual error. The task asks for a rectangle inscribed in a circle, but the model produces a circle inside a square, reversing the intended containment. This highlights a failure to ground symbolic reasoning in geometric constraints, where surface-level textual correctness leads to invalid constructions.

\section{Conclusion}
\label{section:conclusion}

We introduce \textbf{GGBench}, the first benchmark explicitly designed to evaluate \emph{geometric generative reasoning}—the ability of a model to not only understand and reason but also to construct verifiable solutions. Each problem in GGBench integrates natural-language instructions, executable GeoGebra code, and rendered diagrams, forming a tri-modal, interpretable, and fully verifiable testbed.
Our unified evaluation protocol jointly measures textual planning quality, code executability and geometric equivalence, and visual diagram fidelity, providing a holistic view of multimodal reasoning competence. Comprehensive experiments across state-of-the-art UMMs and LLMs/LRMs reveal a consistent gap between end-to-end image generation and reasoning-grounded construction. While UMMs excel at perceptual synthesis, they often fail to satisfy precise geometric constraints; in contrast, code-generating models demonstrate higher logical coherence and geometric correctness but remain limited in visual expressivity. We hope GGBench will serve as a rigorous foundation for future research in grounded, verifiable multimodal intelligence. 
Beyond geometry, our framework points toward a broader paradigm for evaluating generative reasoning—linking \emph{language, logic, and construction}—as a step toward AI systems capable of building as well as understanding.


\clearpage
\newpage
\bibliographystyle{plainnat}
\setcitestyle{numbers}
\bibliography{ref}

\clearpage
\newpage
\beginappendix

\section{Data Curation Prompts}
\label{app:data_curation}

\subsection{LLM-i - Reformulation Suitability Screening}

To operationalize stage~(b) in our pipeline (Figure~\ref{fig:pipeline}), we employ \emph{GPT-5~\cite{OpenAI2025GPT5SystemCard}} as a binary judge that decides whether a raw problem can be \emph{reformulated into a geometric construction task}. The model returns a single token (\texttt{true}/\texttt{false}) without explanations, enabling high-throughput triage before manual verification. Figure~\ref{fig:gpt5_screen_prompt} shows the exact prompt template used in our experiments.

\begin{figure}[h!]
  \centering
  \begin{promptbox}[LLM-i: Reformulation Suitability Judge]
  \textbf{Role.} You are a \emph{Geometric Construction Feasibility Judge}.  
  Given a math problem (text + optional diagram), your task is to \textbf{only decide}
  whether the problem is \textbf{suitable to be reformulated as a geometric construction problem}.  
  \textbf{Output only one word:} \texttt{true} \textbf{or} \texttt{false}. Do not output any explanation, punctuation, or spaces.

  \medskip
  \textbf{Criteria}

  \textit{All} of the following must hold (otherwise output \texttt{false}):
  \begin{itemize}\setlength{\itemsep}{2pt}
    \item \textbf{Geometric objects are explicit:} the statement involves planar geometric objects/relations (points, lines, angles, circles, triangles/polygons, parallels/perpendiculars, angle bisectors, midpoints, centers, tangents, intersections, etc.).
    \item \textbf{Convertible into a construction goal:} there is a clear object/position/figure/relation to be \emph{constructed}, not merely a proof or numeric calculation.
    \item \textbf{Reasonable determinacy:} under common assumptions (not to scale; free choice of references), the target has a clearly attainable solution (unique or finitely many), not severely underdetermined.
  \end{itemize}

  Output \texttt{false} if the problem is purely algebraic/calculus/probability/statistics/equation solving or analytic calculations with no geometric construction goal.

  \medskip
  \textbf{Output requirements (must be strictly followed)}
  \begin{itemize}\setlength{\itemsep}{2pt}
    \item Only output: \texttt{true} or \texttt{false}.
    \item Do not output any other characters, spaces, or line breaks.
  \end{itemize}

  \medskip
  \textbf{Inputs}
  \begin{itemize}\setlength{\itemsep}{2pt}
    \item Problem text: \texttt{\{PROBLEM\_TEXT\}}
    \item Problem diagram: \texttt{\{OPTIONAL\_DIAGRAM\}}
  \end{itemize}
  \end{promptbox}
  \caption{Prompt template used by \textbf{GPT-5~\cite{OpenAI2025GPT5SystemCard}} in pipeline stage~(b) to screen whether a raw problem can be reformulated as a geometric construction task. The template enforces a single-token decision (\texttt{true}/\texttt{false}) with explicit positive criteria and disqualifiers, enabling high-throughput automatic triage prior to human verification.}
  \label{fig:gpt5_screen_prompt}
\end{figure}

\subsection{LLM-ii - Geometric Problem Generation}
\label{app:prompt_llmii}

\begin{figure}[h]
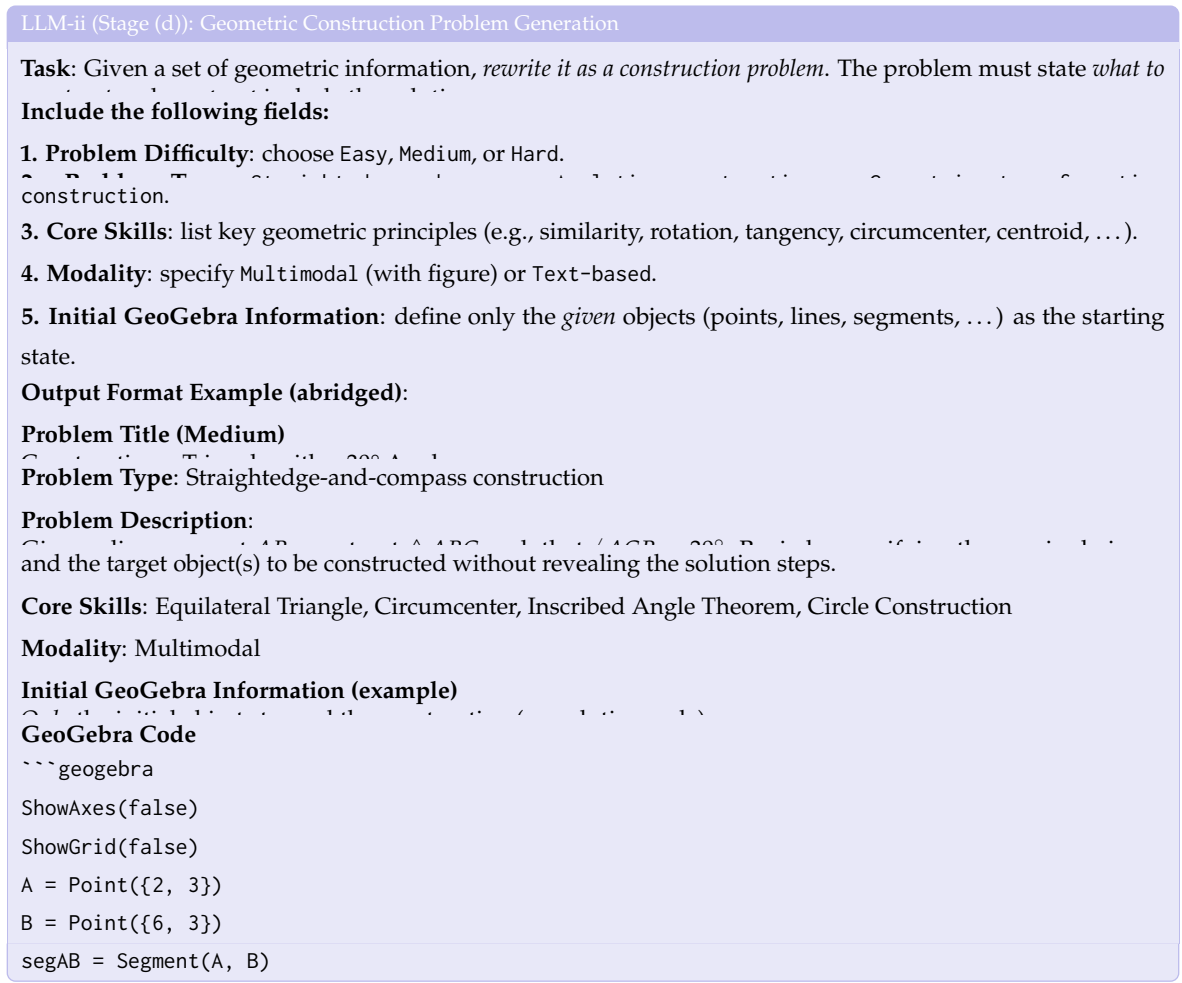

\centering
\begin{promptbox}[LLM-ii (Stage (d)): Geometric Construction Problem Generation]
{\small
\textbf{Task}: Given a set of geometric information, \emph{rewrite it as a construction problem}. The problem must state \emph{what to construct} and must not include the solution.

\textbf{Include the following fields:}

\textbf{1. Problem Difficulty}: choose \texttt{Easy}, \texttt{Medium}, or \texttt{Hard}.\\
\textbf{2. Problem Type}: \texttt{Straightedge-and-compass}, \texttt{Analytic construction}, or \texttt{Geometric transformation construction}.\\
\textbf{3. Core Skills}: list key geometric principles (e.g., similarity, rotation, tangency, circumcenter, centroid, \dots).\\
\textbf{4. Modality}: specify \texttt{Multimodal} (with figure) or \texttt{Text-based}.\\
\textbf{5. Initial GeoGebra Information}: define only the \emph{given} objects (points, lines, segments, \dots) as the starting state.

\textbf{Output Format Example (abridged)}:

\textbf{Problem Title (Medium)}\\
Constructing a Triangle with a 30° Angle

\textbf{Problem Type}: Straightedge-and-compass construction

\textbf{Problem Description}:\\
Given a line segment $AB$, construct $\triangle ABC$ such that $\angle ACB = 30^\circ$. Begin by specifying the required givens and the target object(s) to be constructed without revealing the solution steps.

\textbf{Core Skills}: Equilateral Triangle, Circumcenter, Inscribed Angle Theorem, Circle Construction

\textbf{Modality}: Multimodal

\textbf{Initial GeoGebra Information (example)}\\
\emph{Only} the initial objects to seed the construction (no solution code).

\textbf{GeoGebra Code}
\begin{verbatim}
```geogebra
ShowAxes(false)
ShowGrid(false)
A = Point({2, 3})
B = Point({6, 3})
segAB = Segment(A, B)
\end{verbatim}
} 
\end{promptbox}
\caption{Prompt template for \textbf{LLM-ii} at stage~(d), aligned with the LLM-iii box in layout and syntax.}
\label{fig:llmii_prompt}
\end{figure}

At stage~(d) of the GGBench pipeline (Figure~\ref{fig:pipeline}), \textbf{LLM-ii} rewrites pre-screened geometry questions into formal \emph{geometric construction problems}. 
The model transforms general geometry descriptions into construction-oriented formulations that specify given elements, target relations, and construction goals—without revealing solutions or GeoGebra code. 
It also assigns metadata including difficulty level, construction type, and core geometric skills.
Figure~\ref{fig:llmii_prompt} shows the prompt template used in this stage (abridged with ellipses for brevity).

\subsection{LLM-iii - Answer Generation}
\label{app:prompt_llmiii}

Based on the adapted problems produced in stage~(d), \textbf{LLM-iii} generates the final \emph{geometric construction answer} at stage~(e) (Figure~\ref{fig:pipeline}). The model is instructed to output stepwise reasoning aligned with executable GeoGebra code; per-step snapshots are required for visual traceability. Figure~\ref{fig:llmiii_prompt} shows the prompt template (abridged with ellipses for brevity).

\begin{figure}[h] \centering \begin{promptbox}[LLM-iii (Stage (e)): Geometric Construction \emph{Answer} Generation] \textbf{Task}: Given a geometric construction problem, generate a \textbf{step-by-step construction answer} including: (1) stepwise construction from primitives (points, lines, circles) to the final figure; (2) \emph{valid GeoGebra code for each step}; (3) \emph{reasoning for each step} (e.g., perpendicular bisector, symmetry, tangency, \dots); (4) per-step snapshots showing figure evolution; (5) clear labels/annotations for all objects. \textbf{GeoGebra code requirements} (important): No comments and no blank lines; strictly follow GeoGebra syntax; define objects before use; only official commands; simple, non-self-intersecting polygons; consistent, descriptive names (\texttt{PointA}, \texttt{LineAB}, \dots). \textbf{Constraints \& notes} (abridged): Verify commands exist and are valid; initialize points/lines/angles before dependent objects; avoid CW/CCW mistakes; ensure angle/ratio sources are stated; check intersections and angles; use consistent notation; angles displayed should be integers when required; \dots \textbf{Checklist}: (i) no undefined objects; (ii) no spelling/syntax errors; (iii) strict GeoGebra compliance. If errors exist, revise. \textbf{Output structure}: Numbered steps; each step provides \emph{Method}, \emph{Principle}, and a GeoGebra code block. \emph{Example (abridged)}: \textbf{Step 1: Construct auxiliary circles to locate the circumcenter}\\ \emph{Method}: draw $c_1$ centered at $A$ with radius $AB$, and $c_2$ centered at $B$ with radius $BA$.\\ \emph{Principle}: intersections of $c_1$ and $c_2$ are equidistant from $A$ and $B$. 
\textbf{GeoGebra Code} 
\begin{verbatim}
geogebra
  ShowAxes(false)
  ShowGrid(false)
  A = Point({2, 3})
  B = Point({6, 3})
  c1 = Circle(A, B)
  c2 = Circle(B, A)
\end{verbatim}
Continue stepwise until completion. Provide a final, single, self-contained GeoGebra code block that runs without errors.
\end{promptbox}
\caption{Prompt template for \textbf{LLM-iii} at stage~(e) in Figure~\ref{fig:pipeline}, used to generate the complete \emph{geometric construction answer} (stepwise reasoning aligned with executable GeoGebra). Ellipses (\dots) indicate omitted boilerplate for brevity.}
\vspace{-6mm}
\label{fig:llmiii_prompt}
\end{figure}

\subsection{LLM-iv Prompt for Final Automatic Screening}
\label{app:prompt_llmiv}

At stage~(f) of the GGBench pipeline (Figure~\ref{fig:pipeline}), \textbf{LLM-iv} acts as the final \emph{automatic construction judge}. 
It reviews the generated construction tasks and corresponding solutions produced by earlier stages, verifying whether each sample meets all structural, syntactic, and geometric requirements before inclusion in the final dataset. 
The model outputs a binary score (\texttt{1} = pass, \texttt{0} = fail) together with diagnostic reasoning in JSON format. 
Figure~\ref{fig:llmiv_prompt} presents the abridged prompt template used for this purpose.

\begin{figure}[h]
\centering
\begin{promptbox}[LLM-iv (Stage (f)): Construction Compliance Verification Prompt]
\textbf{Task}: Given a geometric construction task, verify whether a model-generated submission satisfies all structural, syntactic, and field-level requirements. The input includes the construction problem, initial diagram code, and a stepwise solution with corresponding GeoGebra commands.

\textbf{Evaluation requirements}:
(1) \emph{Information extraction}: extract title, difficulty, problem type, description, core skills, modality, initial code block, and construction steps;  
(2) \emph{Problem reasonableness}: target must be clear; constraints must be sufficient; construction must be geometrically feasible;  
(3) \emph{Structural and field compliance}: allowed types/modalities only; English-only text; a single initial GeoGebra code block (no comments or blank lines); no undefined or malformed objects;  
(4) \emph{Stepwise consistency}: each step must explain “what + why”; commands must match text; logical progression must be followed.

\textbf{Scoring rubric}:
(1) reasonable construction and full compliance.  
(2) any mandatory rule violated.

\textbf{Output structure}:  
Provide a brief analysis, followed by a strict binary score:

\textbf{[Scoring Rationale]:} explain format, compliance, and syntax checks (e.g., English-only, valid commands, no undefined objects, no blank lines, consistent use of ShowAxes, ZoomIn, etc.)

\textbf{[Score]:} X point(s)

\textbf{[JSON]:}
\begin{verbatim}
{"answer_score": [[X]]}
\end{verbatim}

End with: “In summary, this submission deserves X point(s).”

\end{promptbox}
\caption{Prompt template for \textbf{LLM-iv} at stage~(f) in Figure~\ref{fig:pipeline}, used to verify structural integrity, field compliance, and command-level correctness in GeoGebra-based construction scripts. The prompt outputs a strict 0/1 score based on format and execution validity.}
\label{fig:llmiv_prompt}
\end{figure}

\subsection{Human Checking \& Filtering Standards}
\label{app:human_filtering}

At stage~(b) of the GGBench pipeline (Figure~\ref{fig:pipeline}), human annotators perform the first-level screening to ensure that all retained samples are geometrically meaningful, visually grounded, and executable when applicable. This step complements the automated filtering conducted by LLM-i and forms a high-quality foundation for downstream prompt adaptation (stage~(d)). The manual filtering process follows three evaluation dimensions as outlined below.

\noindent\textbf{(1) Geometric relevance.}  
Each problem must explicitly present geometric semantics or spatial relationships. Accepted problems typically contain characteristic geometric terminology such as “point,” “line,” “angle,” “circle,” “triangle,” “parallel,” “perpendicular,” “radius,” or “area.” Problems without explicit geometric keywords may still qualify if their content clearly implies geometric reasoning, such as construction, angle relations, or area derivation. In contrast, problems focused purely on algebraic manipulation, logical deduction, or other non-spatial reasoning are excluded.

\noindent\textbf{(2) Image–text correspondence.}  
Each problem must include an accompanying geometric diagram or schematic that accurately reflects the textual conditions and spatial configuration. The image should support reasoning or serve as visual evidence for the described construction. Samples are removed if their images lack geometric characteristics, contradict the textual description, or contain unrelated visual content such as natural photographs, tables, or decorative graphics.

\noindent\textbf{(3) Code executability and consistency.}  
For problems containing executable code (e.g., GeoGebra scripts, or LaTeX drawing commands), annotators manually verify that all code runs correctly in standard environments. The generated output must align with the geometric meaning and textual description. Any instance with syntax errors, runtime failures, or inconsistencies between the produced figure and the problem statement is excluded from the dataset.

This threefold manual screening protocol ensures that all retained items exhibit clear geometric validity, accurate multimodal alignment, and functional code integrity—providing a robust and verifiable foundation for subsequent automated evaluation and dataset finalization.

\subsection{Expert Validation Standards}
\label{app:expert_validation}

At stage~(f) of the GGBench pipeline (Figure~\ref{fig:pipeline}), expert reviewers conduct the final validation to ensure the scientific soundness, structural completeness, and executable reliability of all retained samples. This stage serves as the last human-in-the-loop checkpoint before dataset finalization, guaranteeing that every item meets research-grade quality requirements. The expert validation process is guided by the following five criteria.

\noindent\textbf{(1) Logical rigor of reasoning structure.}  
Experts examine whether each problem demonstrates a complete and coherent reasoning chain that conforms to mathematical logic and geometric principles. Samples with missing premises, invalid inferential steps, or conclusions that cannot be justified by the given conditions are considered invalid.

\noindent\textbf{(2) Consistency and correctness of drawing procedures.}  
Problems involving geometric constructions, function plotting, or visual rendering are manually reviewed to ensure that all drawing instructions and steps follow established geometric conventions. The generated figures must correspond precisely to the textual descriptions. Samples are rejected if inconsistencies occur between text and graphics, if proportions are distorted, or if essential annotations—such as angles, lengths, or intersection points—are missing.

\noindent\textbf{(3) Accuracy and reproducibility of results.}  
Experts verify that each final outcome, including computed values, geometric figures, or rendered visualizations, is both correct and reproducible. This involves cross-checking the stated results with mathematical reasoning and the actual rendering outputs. Any instance of computational error, logical inconsistency, or semantic mismatch with the problem intent is disqualified.

\noindent\textbf{(4) Clarity and formal correctness of presentation.}  
Each problem is reviewed for linguistic clarity, symbolic correctness, and graphical readability. The use of mathematical symbols and notations must follow formal conventions, and the visual layout must support interpretability in research contexts. Problems containing ambiguous expressions, inconsistent symbols, or irregular formatting are revised or excluded.

\noindent\textbf{(5) Global coherence and research alignment.}  
Finally, experts assess whether the overall logic, visual presentation, and code execution results are semantically aligned. Only problems that demonstrate complete coherence across content, structure, and visualization—and that fall within the intended scope of multimodal geometric reasoning—are retained in the finalized GGBench dataset.

This expert validation phase ensures that the final dataset satisfies high standards of mathematical soundness, visual consistency, and executable reliability, enabling robust benchmarking for future research in multimodal geometric reasoning and construction.

\section{Evaluation}

\subsection{Experimental Setups}
\label{app:exp_setup}
To ensure deterministic outputs, we fix the temperature at 0.0 during inferencet that eliminates stochasticity and enforces consistent step-by-step reasoning. All code generation adheres strictly to the GeoGebra command syntax to ensure executability and structural correctness. Inference is parallelized for efficiency. All models are evaluated under the same settings with prompts and pipelines.

\subsection{VLM-T Scoring Prompt}
\label{app:vlm_t}

\autoref{fig:vlmtext_prompt} illustrates the evaluation prompt used to assess VLM-T performance, one of the core dimensions in GGBench. This component focuses on evaluating the textual reasoning in solving geometric construction problems.

\begin{figure}[h]
  \centering
  \begin{promptbox}[LLM-text-score: Evaluation Prompt for VLM-T]
  \sloppy\setlength{\emergencystretch}{1em}

  \textbf{Role}: You are an evaluator of the “text steps” for geometric multimodal constructions.

  \textbf{Input}:  
  \begin{itemize}\setlength{\itemsep}{1pt}
    \item Problem description  
    \item Reference answer: standard construction steps  
    \item Model’s answer: model’s construction steps 
  \end{itemize}

  \textbf{Evaluation Criteria}:
  \begin{itemize}\setlength{\itemsep}{1pt}
    \item \textbf{Completeness}: step order is clear; no skipped logic; dependency relations are explicit  
    \item \textbf{Accuracy}: key geometric operations are correctly described  
    \item \textbf{Geometric Consistency}: operations comply with geometric constraints  
    \item \textbf{Reference Equivalence}: alternative but equivalent strategies are acceptable  
  \end{itemize}

  \textbf{Scoring Rubric (1–5)}:
  \begin{itemize}\setlength{\itemsep}{1pt}
    \item 5: Complete, rigorous, and equivalent to the reference answer  
    \item 4: Minor omissions or differences that do not affect reproducibility  
    \item 3: Most key points covered but with notable omissions or errors  
    \item 2: Numerous logical flaws or incoherent reasoning  
    \item 1: Invalid, unstructured, or irreproducible construction  
  \end{itemize}

  \textbf{Output Requirements}:  
  Return a single Arabic numeral (1–5) only. No text, punctuation, or justification.
  \end{promptbox}
  \caption{Prompt used to compute VLM-T scores. The evaluator assesses textual construction steps on logical, geometric, and structural grounds, returning a scalar score between 1 and 5. }
  \label{fig:vlmtext_prompt}
\end{figure}

The evaluation process compares a model's generated construction steps against expert-written reference steps, emphasizing logical completeness, geometric correctness, and stepwise coherence. A rubric-based scoring from 1 to 5 is adopted, where higher scores indicate precise and faithful reasoning. To maintain scoring consistency, the evaluator is instructed to ignore stylistic variations and accept logically equivalent strategies that achieve the same geometric objectives. The final score serves as a proxy for the model’s ability to translate visual reasoning tasks into accurate symbolic instructions, and directly contributes to the VLM-T metric reported in \autoref{tab:main_results}.

\subsection{VLM-I-Res Scoring Prompt}
\label{app:vlm_res}

\autoref{fig:vlmimage_prompt} shows the evaluation template used to obtain VLM-I-Res scores in GGBench.

\begin{figure}[!ht]
  \centering
  \vspace{2mm}
  \begin{promptbox}[LLM-image-score: Evaluation Prompt for VLM-I-Res]
  \textbf{Task}: You are a geometric image consistency evaluator.

  \textbf{Input}: 
  \begin{itemize}\setlength{\itemsep}{1pt}
    \item Reference answer image
    \item Model’s final image
  \end{itemize}

  \textbf{Evaluation Criteria}:
  \begin{itemize}\setlength{\itemsep}{1pt}
    \item Element completeness: whether the key geometric elements required by the problem are included (points, line segments, circles/auxiliary circles, points of tangency, tangents, annotations/labels, etc.).
    \item Topology \& constraints: whether relative positions and geometric relationships (perpendicular, parallel, tangency, concyclicity/collinearity, angle/proportional relationships, etc.) are correct.
    \item Visual tolerance: do not penalize non-critical style differences such as line width, color, fonts; do not penalize translation/rotation/uniform scaling/mirroring (similarity transforms) unless explicitly forbidden by the problem.
    \item Overall judgment: prioritize the correctness of geometric topology and constraints; LPIPS is only a reference for perceptual similarity—if there is a conflict, geometric consistency takes precedence.
  \end{itemize}

  \textbf{Scoring Rubric (1–5)}:
  \begin{itemize}\setlength{\itemsep}{1pt}
    \item 5: Elements, topology, and constraints are highly consistent
    \item 4: Basically consistent, only slight positional/style deviations
    \item 3: Mostly consistent; elements are complete but there are several minor errors/deviations
    \item 2: Only partially consistent; key elements are missing or constraint errors are obvious
    \item 1: Inconsistent with the problem intent or missing key elements (such as circles/auxiliary circles/points of tangency/tangents), or the image is unusable
  \end{itemize}

  \textbf{Boundary Handling}:
  \begin{itemize}\setlength{\itemsep}{1pt}
    \item Missing/corrupted/unreadable image → 1
    \item Text only with no image → 1
  \end{itemize}

  \textbf{Output Requirements}: Output only a single Arabic numeral (1–5) as the final result, with no other text, punctuation, or spaces.

  \end{promptbox}
  \caption{Prompt used to compute the VLM–I-Res score.
  The VLM judge compares the model’s final rendered diagram against the reference image 
  and assigns a 1–5 rating based on geometric consistency and constraint satisfaction.}
  \label{fig:vlmimage_prompt}
\end{figure}

This prompt guides GPT-4o to act as an impartial geometric evaluator that judges the consistency between a model’s \emph{final rendered diagram} and the \emph{reference solution}.
The design emphasizes geometric reasoning over superficial appearance, ensuring that perceptual similarity does not override structural correctness.


\begin{figure}[h]
  \centering
  \begin{promptbox}[VLM-judge: Prompt for Step Accuracy Evaluation (Step)]

  \textbf{Task}: Given a geometric construction problem and a rendered long image showing the step-by-step solution, judge whether each step image strictly matches its corresponding textual instruction in terms of naming, geometric structure, and positioning.

  \textbf{Scoring Rules}:
  \begin{itemize}\setlength{\itemsep}{1pt}
    \item \textbf{5 points}: The image and text match perfectly, including naming, topological relationships, and quantity, with no extraneous or missing elements.
    \item \textbf{4 points}: The image and text are mostly consistent, with only very slight visual deviations that do not affect understanding (e.g., a point label is slightly offset but named correctly).
    \item \textbf{3 points}: The image meets the main structural requirements but has more than one minor error (e.g., confusing names, incorrect position of auxiliary points).
    \item \textbf{2 points}: The image and text are difficult to correspond one-to-one; there are critical naming errors, connection errors, or missing elements.
    \item \textbf{1 point}: The overall structure of the image is incorrect, naming is chaotic, it is completely disconnected from the text, or the image is irrelevant/completely incorrect.
  \end{itemize}

  \textbf{Mandatory Clauses}:
  \begin{itemize}\setlength{\itemsep}{1pt}
    \item If naming errors or omissions make it impossible to uniquely identify the target point/line, the score must not exceed \textbf{1 point}.
    \item If the text requires connecting a specific pair of points, but the endpoints in the image cannot be reasonably matched (wrong position \textbf{and} wrong name), score \textbf{1 point}.
    \item If a name is slightly misplaced but still identifiable in context, a score of \textbf{2 points} may be given.
  \end{itemize}

  \end{promptbox}
  \caption{Prompt used for Step Accuracy (Step) evaluation by the judge model. This criterion assesses alignment between visual steps and symbolic instructions.}
  \label{fig:judge_prompt_step}
\end{figure}

\subsection{VLM-I-Mid Scoring Prompt}
\label{app:vlm_mid}

To evaluate the consistency and correctness of intermediate construction steps, we employ two complementary visual-judging criteria: 
\textbf{Step Accuracy (Step)} and \textbf{Process Consistency (Consist.)}.
Each is independently scored by a frozen VLM (GPT-4o) using fixed prompts.
The final \texttt{VLM–I\textsubscript{Mid}} score is computed as the mean of the two ratings and rescaled to $[0,100]$.

\vspace{-4mm}

\paragraph{Prompt for Step Accuracy}
\autoref{fig:judge_prompt_step} presents the evaluation prompt used for measuring Step Accuracy. This metric quantifies the degree of alignment between each textual instruction and its corresponding geometric subfigure in the rendered construction sequence. The prompt guides the judge model to assess whether visual elements—including object naming, positional structure, and geometric relationships—faithfully match the symbolic description at each reasoning step. It ensures that the geometric process remains verifiable and syntactically grounded at the step level.

\begin{figure}[ht]
  \centering
  \begin{promptbox}[VLM-judge: Prompt for Process Consistency Evaluation (Consist.)]

  \textbf{Task}: Evaluate whether each step reasonably builds upon the previous step's figure in a cumulative manner, avoiding skipped steps, missing procedures, or logical gaps.

  \textbf{Scoring Rules}:
  \begin{itemize}\setlength{\itemsep}{1pt}
    \item \textbf{5 points}: Completely based on the previous step, structure is fully inherited, and the new construction is correct and complete.
    \item \textbf{4 points}: Mostly coherent, with only minor issues in inheritance clarity or naming deviations.
    \item \textbf{3 points}: The structural relationship between steps is roughly preserved, but there are missing key construction marks, logical skips, or incorrect step order.
    \item \textbf{2 points}: Most constructions are not inherited, or construction continues on an incorrect structure; the logical chain is unclear.
    \item \textbf{1 point}: Steps are severely incoherent, as if they are independent drawings, with no evolutionary relationship visible, or the figures in each step are completely unrelated, forming a broken chain.
  \end{itemize}

  \textbf{Mandatory Clauses}:
  \begin{itemize}\setlength{\itemsep}{1pt}
    \item If the long image contains only one single figure, the score for this dimension must not exceed \textbf{1 point}.
    \item If intermediate figures are missing or replaced by error messages, the score must be \textbf{1 point}.
  \end{itemize}

  \end{promptbox}
  \vspace{-2mm}
  \caption{Prompt used for Process Consistency (Consist.) evaluation by the judge model. This criterion checks visual and logical coherence across sequential construction steps.}
  \vspace{-4mm}
  \label{fig:judge_prompt_consist}
\end{figure}

\paragraph{Prompt for Process Consistency}
\autoref{fig:judge_prompt_consist} illustrates the prompt designed for evaluating Process Consistency. This indicator measures whether each step in a geometric construction logically inherits and extends the previous figure, maintaining coherent spatial and structural progression. The prompt enforces sequential reasoning validity, penalizing disjointed or skipped steps, reflecting the model’s capacity for stable geometric evolution.

\subsection{Prompt for Model Inference}
\label{app:prompt_infer}

To ensure fair and reproducible evaluation across heterogeneous model families, we design two complementary prompt templates corresponding to the two evaluation tracks in GGBench: 
(1) \textbf{LLM/LRM code–based prompting} for models that generate explicit \textit{GeoGebra programs}, and 
(2) \textbf{UMM image–based prompting} for models that directly produce diagrams or visual construction steps. 
Both prompts are carefully standardized to (i) unify task semantics, (ii) minimize ambiguity in geometric intent, and (iii) encourage interpretable, stepwise reasoning.

\begin{figure}[h]
\centering
\begin{promptbox}[Model Inference Prompt for LLMs/LRMs]
{\small
\textbf{Task}:\\
According to the given drawing topic and image, create a construction using GeoGebra language. 
If the problem requires a ruler-and-compass construction, retain all auxiliary traces. 
Output step-by-step drawing instructions, geometric principles, and executable GeoGebra code.

\vspace{4pt}
\textbf{Code Generation Rules:}
\begin{enumerate}
  \item Follow GeoGebra syntax strictly (capitalize commands; no underscores).
  \item Use \texttt{Point(\{x, y\})} for points and \texttt{Vector((x, y))} or \texttt{Vector(Point1, Point2)} for vectors.
  \item Retain all auxiliary elements in the final figure.
  \item Use a single coordinate system; no comments or blank lines.
  \item Only static drawings are allowed (no animations).
\end{enumerate}

\vspace{4pt}
\textbf{Output Example:}\\
\emph{Problem Title:} Construct the incircle of triangle $ABC$.\\
Each step includes \textbf{Construction}, \textbf{Principle}, and \textbf{GeoGebra Code} sections.

\textbf{GeoGebra Code (abridged):}
\begin{verbatim}
ShowAxes(false)
ShowGrid(false)
A = Point({1, 5})
B = Point({0, 1})
C = Point({7, 2})
...
bisA = AngleBisector(C, A, B)
bisB = AngleBisector(A, B, C)
I = Intersect(bisA, bisB)
Circle(I, A)
...
ZoomIn(0, 0, 8, 6)
\end{verbatim}
}
\end{promptbox}
\caption{Prompt used for LLMs/LRMs.}
\label{fig:llminfer_prompt}
\end{figure}

Figure~\ref{fig:llminfer_prompt} shows the template used to elicit explicit \emph{text–to–code} reasoning.
Each instance provides a natural–language construction topic and, when available, a reference diagram. Strict code–generation rules are enforced to guarantee syntactic validity and deterministic rendering—e.g., capitalized command names, no underscores, static drawings only, and a single coordinate frame.
For ruler–and–compass problems, auxiliary traces must be preserved to make the reasoning chain fully inspectable.
An abridged output example is provided in the prompt box of Fig.~\ref{fig:llminfer_prompt}.
This design encourages models to \emph{reason, formalize, and construct} in a single pipeline, directly linking linguistic reasoning to executable geometry.

\begin{figure}[t]
  \centering
  \small
  \begin{promptbox}[Model Inference Prompt for UMMs]
  \textbf{Task}: 
  
  According to the given drawing topic, create a drawing that meets the requirements. If the problem requires the use of a ruler and compass construction, please retain the traces of ruler and compass construction. Finally, only output the drawing steps.

  \textbf{Drawing Step Generation Rules:}
  \begin{enumerate}
  \item Strictly follow the description of the drawing steps and clearly explain each step.
  \item The output should only include detailed step descriptions, without any code.
  \item The steps should describe the drawing process, ensuring that each description leads to the creation of the corresponding image.
  \item The steps should be clear and concise, avoiding dynamic effects or code.
  \end{enumerate}

  \vspace{4pt}
  \textbf{Output Example:}\\
  \emph{Problem Title:} Given two circles of different radii $c_1$ (center $O_1$, radius $r_1$) and $c_2$ (center $O_2$, radius $r_2$), which do not intersect. Using only ruler and compass construction, draw the two external tangents to these two circles.

  \textbf{Drawing (abridged):}
  \begin{enumerate}
      \item Construct the angle bisector of ( $\angle$ BAC ).
      \item Construct the angle bisector of ( $\angle$ ABC ).
  \end{enumerate}
   \textbf{Principle:}
   The incenter of a triangle (the center of the incircle) is the intersection of its three angle bisectors. According to the angle bisector theorem, any point on an angle bisector is equidistant from the two sides of the angle. Therefore, the intersection of the two angle bisectors will be equidistant from the three sides of the triangle.

  \end{promptbox}
  \caption{Prompt used for UMMs (e.g., Janus, Bagel, Qwen-Image, Nano Banana, Seeddream).}
  \vspace{-2mm}
  \label{fig:imagegen_prompt}
\end{figure}

Figure~\ref{fig:imagegen_prompt} illustrates the template for end-to-end multimodal systems that generate drawings directly from text.
These models receive the same construction topic but are instructed to produce only natural–language descriptions of drawing steps without emitting code.  
The rules emphasize clarity and procedural completeness: each step must describe a concrete geometric operation that would yield the corresponding figure; dynamic or stylistic embellishments are prohibited.
If the problem involves ruler-and-compass construction, all traces should be retained in the generated image. This prompt design isolates the model’s visual \emph{generative reasoning} ability—its capacity to transform textual geometric constraints into coherent visual sequences—without relying on symbolic code execution.


\subsection{Human Evaluation}
\label{app:human_eval}

\begin{figure}[t]
  \centering
  \includegraphics[width=0.8\linewidth]{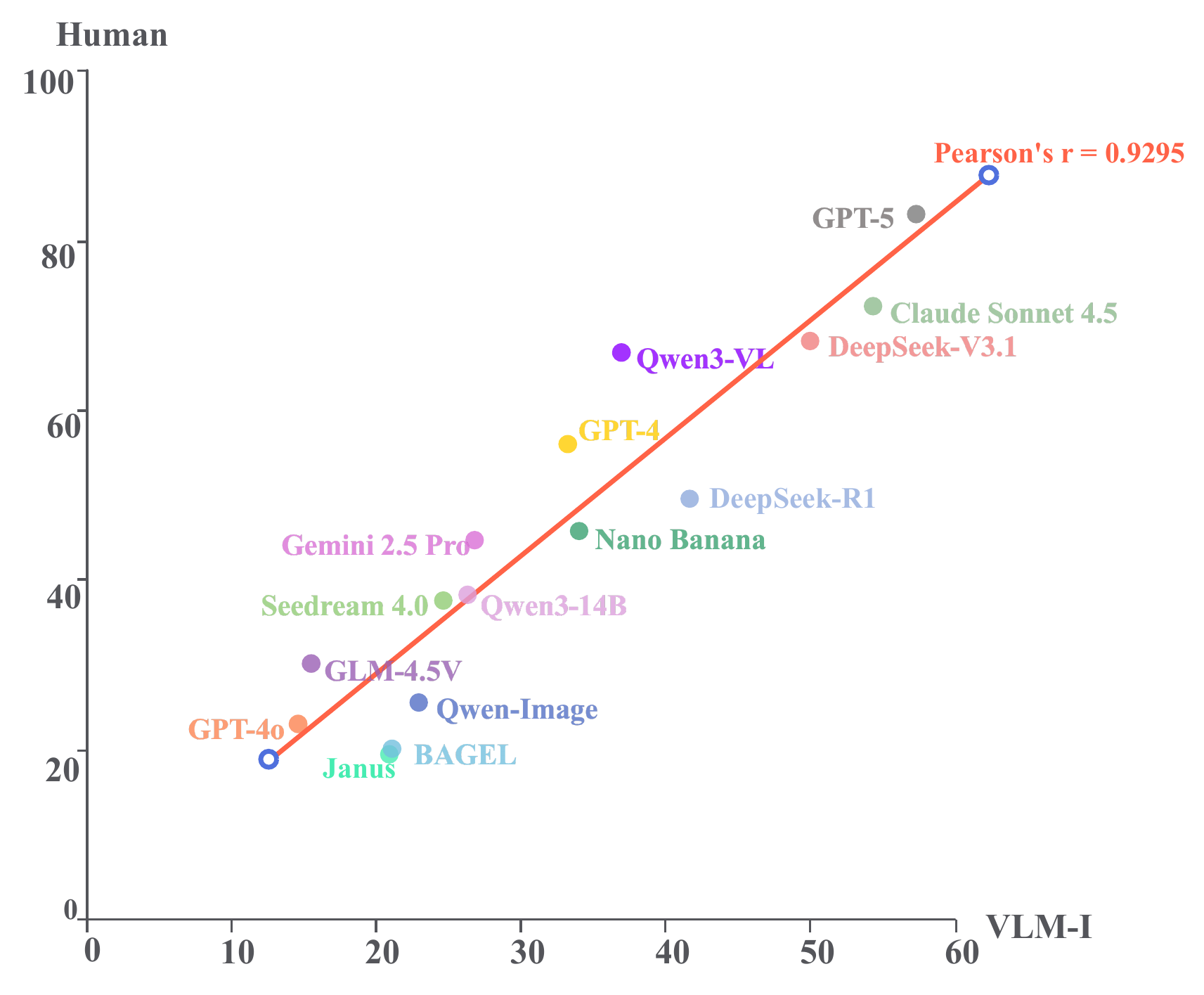}
  \caption{Correlation between VLM-I and human evaluation. }
  \label{fig:human_eval_scatter}
\end{figure}
To verify the reliability and consistency of our automatic evaluation metric (VLM-I), we conducted a systematic human evaluation involving 3 domain experts with backgrounds in mathematics education and geometric modeling. For each model, 100 samples were randomly selected and independently scored to ensure representativeness and fairness.

Prior to the evaluation, all experts underwent standardized training to ensure familiarity with the assessment protocol, scoring rubric, and exemplar responses. A double-blind review mechanism was adopted to minimize subjective bias:
\begin{itemize}
    \item Each sample was independently evaluated by two experts;
    \item If the score difference exceeded 1 point, a third expert acted as arbiter;
    \item The final score was the average of the three expert ratings.
\end{itemize}

\paragraph{Evaluation Criteria.} Experts evaluated each model’s generation by jointly considering its textual construction steps and the rendered figure. Scores were assigned based on three criteria: logical soundness of the geometric reasoning, accuracy of the drawing process, and overall consistency between text and image. A 1–5 scale was used:

\begin{itemize}
    \item \textbf{5 – Perfect Match}: Both the textual steps and diagram are complete, logically sound, and closely match the reference. All geometric primitives (e.g., circles, auxiliary lines, intersections, tangents) are correctly included and rendered.
    \item \textbf{4 – Mostly Correct}: Minor deviations in position, style, or step detail, but the core reasoning and visual fidelity are intact.
    \item \textbf{3 – Partially Correct}: The key reasoning path is preserved, but notable omissions or inconsistencies exist (e.g., missing constructions, misaligned diagrams, or loose logical steps).
    \item \textbf{2 – Major Errors}: Either the text or diagram deviates significantly from the problem intent, omitting several key steps or introducing logical flaws.
    \item \textbf{1 – Invalid Result}: Severe inconsistency or mismatch between text and diagram, with critical geometric elements missing or misinterpreted.
\end{itemize}

\paragraph{Correlation Analysis.} To assess alignment between human judgments and automated scores, we computed the Pearson correlation coefficient ($r$) between VLM-I scores and average human ratings. As shown in Figure~\ref{fig:human_eval_scatter}, the correlation is exceptionally high ($r = 0.9295$), confirming that our automated metric closely reflects human-perceived quality and faithfully captures model performance trends.

\section{Case Study}

\subsection{Examples across Difficulty Levels}

\begin{figure}[ht]
  \centering
  \includegraphics[width=\linewidth]{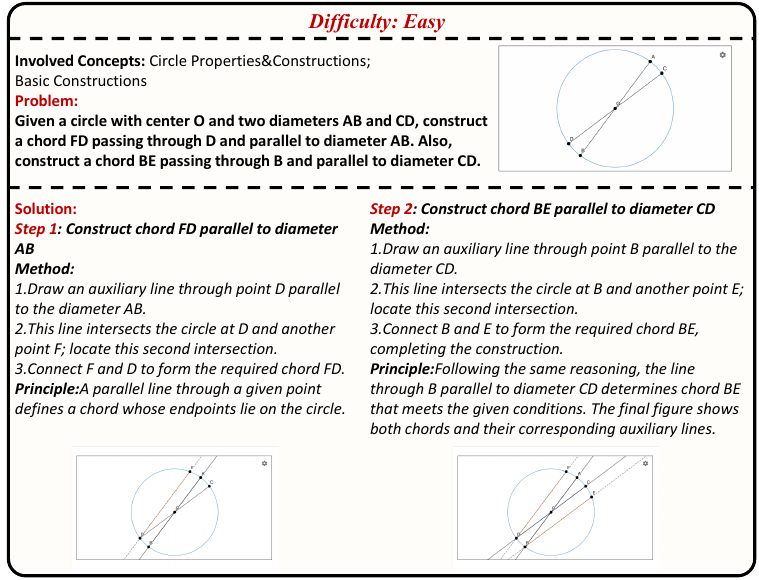}
  \caption{Example of an easy-level task.}
  \label{fig:example_easy}
\end{figure}

\begin{figure}[ht]
  \centering
  \includegraphics[width=\linewidth]{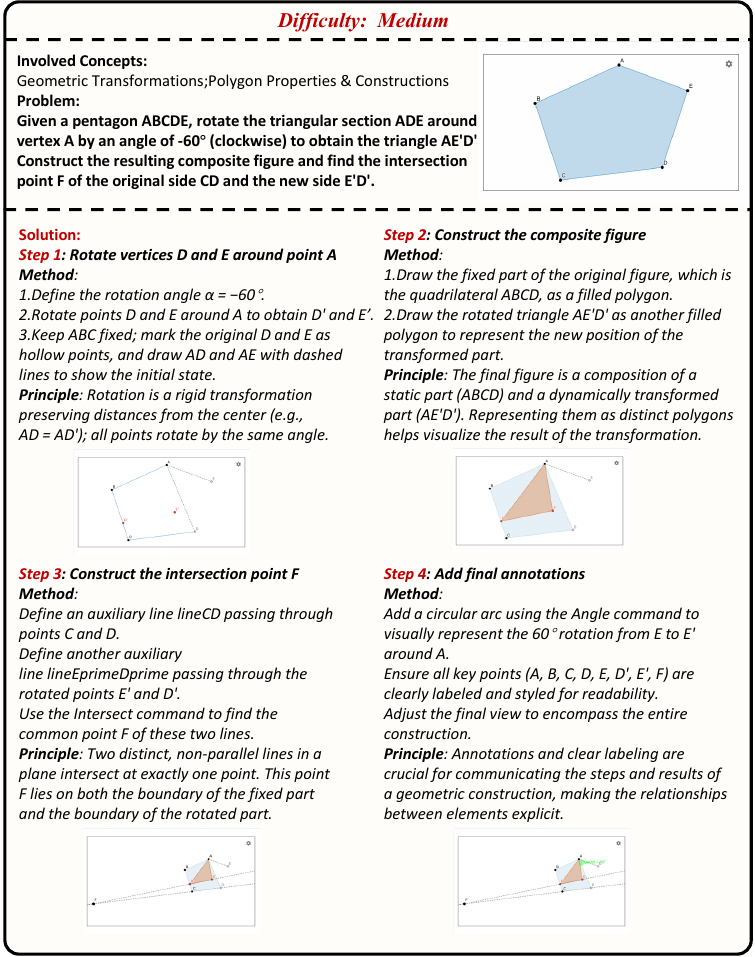}
  \caption{Example of a medium-level task.}
  \label{fig:example_medium}
\end{figure}

\begin{figure*}[ht]
  \centering
  \includegraphics[width=0.98\linewidth]{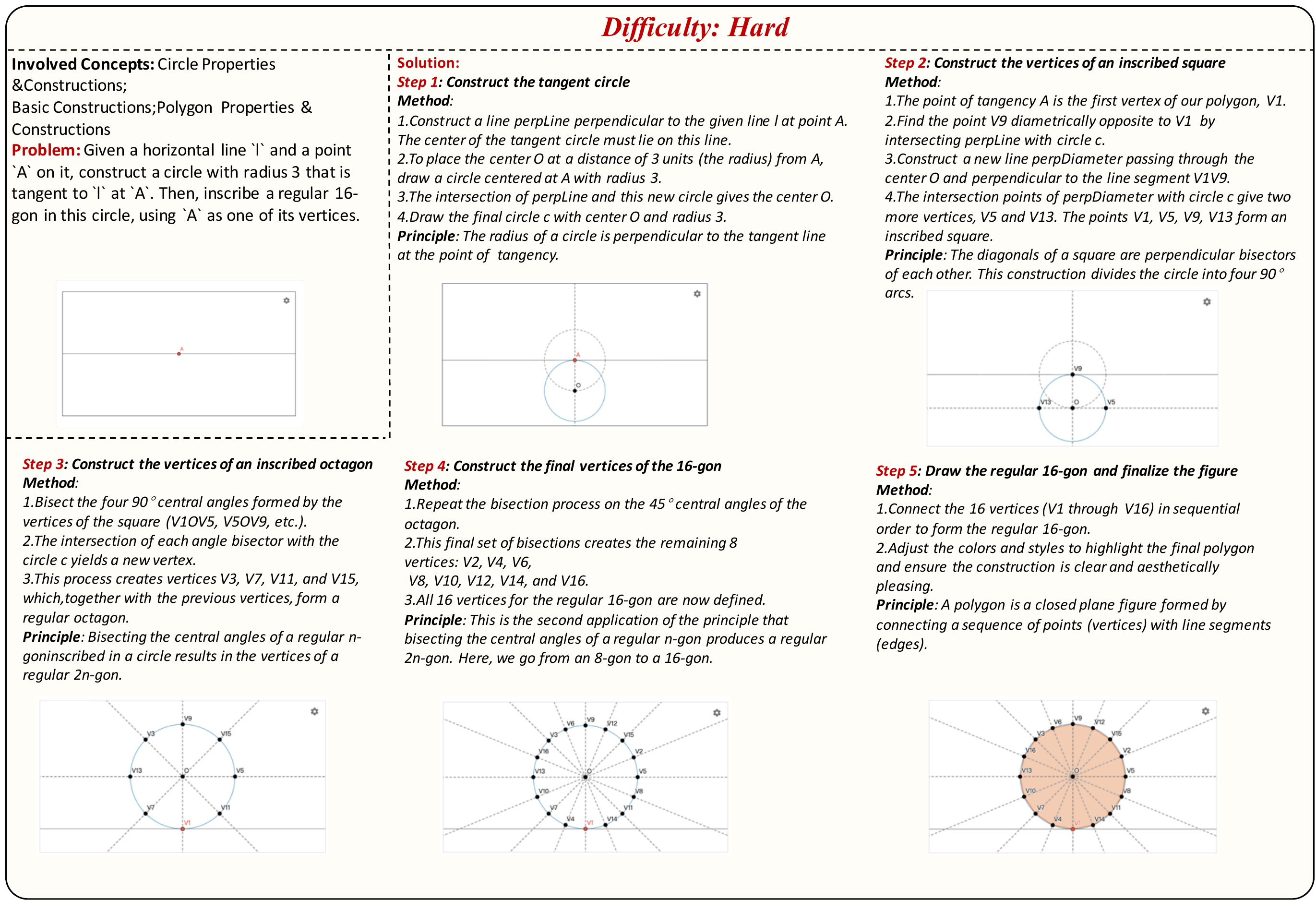}
  \caption{Example of a hard-level task.}
  \label{fig:example_hard}
\end{figure*}

To illustrate the range of geometric reasoning tasks in GGBench, we present three representative examples drawn from the benchmark, each aligned with a specific difficulty level. Figures~\ref{fig:example_easy}–\ref{fig:example_hard} visualize problems designed to probe distinct levels of spatial abstraction, procedural depth, and symbolic control.

Figure~\ref{fig:example_easy} corresponds to an entry-level task grounded in elementary circle geometry. The problem involves constructing chords that pass through given points while remaining parallel to two specified diameters. Despite its simplicity, solving the problem requires maintaining directional consistency, identifying appropriate auxiliary constructions, and accurately positioning endpoints within a bounded coordinate frame. It targets the model’s ability to handle localized spatial constraints and execute basic primitives such as `Segment`, `ParallelLine`, and `Point`. 

Figure~\ref{fig:example_medium} illustrates a transformation-based task involving a regular pentagon and an inscribed triangle. The objective is to rotate the triangle about one vertex and identify the resulting intersections with the original figure. This setup tests the model’s understanding of rigid motions, label consistency, and intersection logic. It requires chaining multiple dependencies while preserving geometric invariants such as distance and angle, and it penalizes any deviation from transformation semantics or object reuse policies.

Figure~\ref{fig:example_hard} represents a high-complexity construction spanning multiple geometric phases. Starting from a tangent circle, the problem unfolds into a sequence of regular polygon constructions—inscribing a square, then an octagon, followed by a 16-gon. The solution involves hierarchical decomposition, recursive angle bisection, and repeated application of symmetric placement rules. Success depends on long-horizon reasoning, internal consistency across stages, and robustness in symbolic command generation under rigid geometric constraints.




\end{document}